\documentclass[journal, compsoc, twoside]{IEEEtran}
\usepackage[T1]{fontenc}
\ifCLASSINFOpdf

\else
\fi

\usepackage[pdftex]{graphicx}
\usepackage{graphicx}
\usepackage{subcaption}
\usepackage{float}
\usepackage{amsmath,amsthm}
\usepackage{optidef}
\usepackage[cmintegrals]{newtxmath}
\usepackage{algorithmic}
\usepackage{xcolor}
\usepackage[vlined,ruled,commentsnumbered,linesnumbered]{algorithm2e}
\usepackage[pdftex, colorlinks, citecolor = blue, anchorcolor = red]{hyperref} 
\newtheorem{rem}{Remark}

\begin{document}

\title{Decentralized Source Localization without Sensor Parameters in Wireless Sensor Networks}


\author{Akram Hussain and Yuan Luo
\thanks{This work is supported by National Natural Science Foundation of China under Grant 61871264.}
\thanks{The authors are with the Department of Computer Science and Engineering, Shanghai Jiao Tong University, Shanghai, China
(e-mail: akram.hussain@iba-suk.edu.pk ; yuanluo@sjtu.edu.cn).}
\thanks{(Corresponding author: Yuan Luo)}
}


\IEEEtitleabstractindextext{
\begin{abstract}
This paper studies the source (event) localization problem in decentralized wireless sensor networks (WSNs) under the fault model without knowing the sensor parameters. Event localizations have many applications such as localizing intruders, Wifi hotspots and users, and faults in power systems. Previous studies assume the true knowledge (or good estimates) of sensor parameters (e.g., fault model probability or Region of Influence (ROI) of the source) for source localization. However, we propose two methods to estimate the source location in this paper under the fault model: hitting set approach and feature selection method, which only utilize the noisy data set at the fusion center for estimation of the source location without knowing the sensor parameters. The proposed methods have been shown to localize the source effectively. We also study the lower bound on the sample complexity requirement for hitting set method. These methods have also been extended for multiple sources localizations. In addition, we modify the proposed feature selection approach to use maximum likelihood. Finally, extensive simulations are carried out for different settings (i.e., the number of sensor nodes and sample complexity) to validate our proposed methods in comparison to centroid, maximum likelihood, FTML, SNAP estimators.
\end{abstract}
}
\maketitle

\IEEEpeerreviewmaketitle

\section{Introduction}
\IEEEPARstart{W}{ireless} sensor networks (WSNs) are being utilized in a variety of applications such as surveillance, health monitoring, environmental monitoring, and industrial monitoring purposes \cite{Kandris}. One of the main reasons for such diverse applications of WSNs is the low cost because usually low cost sensor nodes are used which are densely deployed (more sensors in an area) in the area being monitored. In a decentralized system, these sensor nodes generally take measurements and forward their decision to the fusion center (FC) which performs estimation or learning tasks, depending on the goal.  Moreover, sensor nodes and a fusion center are connected through the coherent orthogonal multiple access channel (MAC) \cite{Xiao}.

The general requirements for sensor nodes are simplicity and low cost, hence, these devices have limited resources (power, memory, and processing/communication capabilities) \cite {MP}, \cite{Liu}. The simple nature of sensor nodes  has some associated disadvantages such as extremely vulnerable to faults which occur due to noise, energy depletion, environmental harsh conditions of operation, attacks, and software problems.

In this paper, we focus on the decentralized source (event) localization under the fault model in which decisions of sensor nodes get reversed. Event localizations have many applications such as localizing intruder, pollutant sources like biological and chemical weapons, enemies positions in combat monitoring \cite{Kandris}, localizing Wifi hotspots and users \cite{Kumar}, \cite{SHe}, localizing drones in urban environment \cite{GYang}, person in indoor and outdoor environment \cite{Hillyard}, \cite{JWang}, and faults in power systems \cite{Kim}. We consider wireless sensor networks for source (event) localization, where sensor nodes monitor the source, and the data set is generated when sensor nodes report their decisions to the fusion center.
Sensors' decisions are binary after comparing their (real) measurements, taking from the source, to some threshold. Some advantages of using binary decisions are as follows \cite{MP}, \cite{Michaelides}:
\begin{itemize}
  \item Binary decisions are simple problems for sensor nodes.
  \item Binary decisions conserve energy because of low bandwidth requirement for communications.
  \item Binary decisions are less sensitive to calibration mismatches and varying sensor sensitivities.
\end{itemize}

Due to the fact that sensor nodes are extremely vulnerable to faults, wireless sensor networks based event (source) localization from binary data have been studied under the fault model in \cite{MP}, \cite{Michaelides}, \cite{Laoudias}, \cite{Xu}, \cite{Yu}. Michaelides and Panayiotou propose the fault tolerant maximum likelihood estimation \cite{MP} by taking into account the fault probability of sensors. Moreover, they also propose an algorithm, SNAP \cite{Michaelides}, which is the simple maximum likelihood algorithm to reduce the effects of faulty sensors, e.g., false positives and false negatives. Multiple event sources localization and tracking are performed in \cite{Laoudias} under the sensor fault model, and they use the distributed version of SNAP for source localization. Furthermore, Xu \textit{et al.} \cite{Xu} modify the distributed SNAP algorithm for multiple sources localization by introducing trust index concept where potential faulty sensor nodes are assigned less weight for future measurements.

Multiple sources localization problem has also been considered in \cite{Sheng}-\cite{Lu} in addition of \cite{Laoudias} and \cite{Xu}. Sheng and Hu \cite{Sheng} propose the maximum likelihood (ML) estimation method which is nonlinear in nature. They propose projection method and expectation maximization like method to solve the nonlinear ML. Multiple disjoint moving sources are localized using time difference of arrival and frequency difference of arrival measurements under the uncertainties of sensor nodes positions and velocities in \cite{Sun}. Xin \textit{et al.} \cite{Xin} use iterative and sequential technique based on compressed sensing to localize multiple sources where the number of sources is unknown. The proposed technique is power efficient in which some sensor nodes get activated for measurements depending on the their distances from the sources. Lu \textit{et al.} propose the expectation maximization algorithm for multiple sources localization under the fault model where the sources experience nonuniform noises \cite{Lu}.

Source localization under noisy communication between sensor nodes and the fusion center and Byzantine attack model: also known as data falsification attack in which sensor nodes send false data to the fusion center, is considered in \cite{Aditya}, \cite{Wei}. In both of these papers, coding technique is used to deal with the Byzantine attack. However, \cite{Aditya} model the source localization problem as an iterative classification problem while \cite{Wei} consider the adaptive local threshold design based on the work of \cite{Aditya} for each sensor node such that the misclassification probability is minimized.

In addition, source localization is performed using real measurements as well from sensor nodes in distributed and centralized environment. Since the maximum likelihood estimation from real measurements is nonconvex, computationally expensive, and difficult to implement in distributive environment, some alternative or modified algorithms are proposed in \cite{Kumar}, \cite{Shi}-\cite{CZhang} for distributed and centralized environment. Heydari \textit{et al.} estimate the source location based on received signal strength difference (RSSD). To optimize the source localization performance, Fisher information matrix is derived which is the function of relative sensor source geometry \cite{Heydari}.

Wireless sensor networks have also been investigated for event detection problem \cite{Liu}-\cite{Ahmed}. Liu \textit{et al.} \cite{Liu} consider the event detection problem under the fault model, and by taking into account the spatiotemporal properties of dense wireless sensor network, they propose two-stage algorithm for event detection under the fault model. Moreover, they also observe that the event detection accuracy degrades rapidly when the sensors' fault probability reaches a critical value. In \cite{Wang}, Bayesian network is developed using the correlation between observed attributes in addition of spatiotemporal characteristics for event detection. Chen \textit{et al.} consider the event detection problem in wireless sensor network by taking into account the spatiotemporal properties \cite{Chen}, where they model the spatial and temporal properties using Markov random field and Markov chain, respectively. In \cite{Krishna}, the delay tolerant event detection algorithm is proposed where the delay is assigned to each sensor by the fusion center based on the distance of each sensor node from it. The advantage of delay tolerant scheme is that the communication congestion, hence, the life time of sensor network can be improved. Furthermore, they also propose a greedy algorithm to search the optimal path to send the sensors' measurements to the fusion center. Ould-Ahmed-Vall \textit{et al.} \cite{Ahmed} consider the distributed event detection problem for heterogeneous sensor network, where each sensor can become faulty with different probability. By utilizing the spatial correlation among sensor nodes, they derive the fault tolerant estimator which is optimal under MAP criteria.

In this paper, we focus on energy based approaches for source localization, also known as received signal strength (RSS) methods which employ maximum likelihood or least squares for source localization. Energy based approaches employ event signal strength sensor measurements to estimate source location because it is more suitable for large scale WSN in the sense that the precise synchronization among sensor nodes is not required \cite{Xu}, \cite{Kumar}.

The main contributions of this paper are as follows.
\begin{itemize}
  \item \textbf{Hitting Set Approach on Noisy Data Set:} First of all, we show in Lemma 1 that Region of Influence (ROI) of the source is equivalent to its neighborhood. The neighborhood is defined as the set of sensor nodes for which there exist edges between the source and the sensor nodes, and an edge will exist between the source and a sensor node if the sensor node is located inside the ROI of the source. We propose to use \textbf{the hitting set approach} to recover the source neighborhood under the fault model. We show in Theorem 1 that source neighborhood can be recovered as the sample size goes to infinite. The advantage of the hitting set approach that it works with the data set without any knowledge of fault probability or region of influence of the source.
  \item \textbf{Sample Complexity Bound for Hitting Set Approach:} In addition, we lower bound the sample complexity (number of samples) requirement in Proposition 1 for hitting set approach such that estimated source neighborhood is equal to true neighborhood with high probability.
  \item \textbf{Feature Selection Approach on Noisy Data Set:} We also propose an algorithm (Algorithm 1) based on feature selection method to recover the source neighborhood, hence, the source location. We also prove in Theorem 2 that Algorithm 1 provides the feasible solution and it also reduces the impact of faulty sensor nodes by using only the noisy data set.
  \item \textbf{Extensions of Hitting Set and Feature Selection Approaches for Multiple Sources:} We propose the extensions of our proposed solutions for multiple sources localization problem. In addition, we propose an iterative algorithm (Algorithm 2) to estimate the sources locations iteratively with the estimations of clusters of sensor nodes.
  \item \textbf{Extensive Experiments:} We validate our proposed solutions through simulations, and show that the proposed methods achieve much better performances in comparison to maximum likelihood and centroid estimators under the fault model. In addition, the proposed modified feature selection approach is shown to perform comparable to FTML and better than SNAP algorithm with small fault model probability when perfect and imperfect knowledge of fault model probability and ROI are available for FTML and SNAP algorithm, respectively.
\end{itemize}

The paper is organized as follows. In Section 2, we make some assumptions, describe the measurement and fault models, and review some source localization estimators. We also formulate the problem statement, and review some definitions there. Main results of this paper are presented in Section 3 and Section 4 for a single source localization and multiple sources localization, respectively. In Section 5, extensive simulation results are presented, and the paper is concluded with Section 6.

\section{Preliminaries and Problem Statement}
In this section, we define the observation model for energy based approaches, review some binary estimators and their problems, and state the fault model and its connection to communication noise. Moreover, we formulate the problem statement.

\subsection{Assumptions} We make the following standard assumptions which are also used in the following papers \cite{MP}-\cite{Laoudias}, \cite{Xu}.
\begin{enumerate}
  \item We uniformly spread $N$ sensor nodes over the rectangular area $A$ where the source may exist. Moreover, the sensor nodes are static, and their positions are known, denoted by $(x_n, y_n)$ for $n = 1, 2, ..., N$.
  \item A source $s$ is located uniformly inside the area $A$, and its location is denoted by $(x_s, y_s)$ which will be estimated in this paper.
  \item The signal is emitted by source $s$, which propagates uniformly in all directions. Moreover, only the distance factor attenuates the emitted signal.
\end{enumerate}

\subsection{Observation Model}
Let the emitted source signal be $c$ which is emitted by the source at its location $(x_s, y_s)$. As it propagates in the environment away from the source, the signal gets attenuated inversely proportional to the distance from the source raised to some power $\alpha \in \mathcal{R}^+$. The constant $\alpha$ is environment dependent.

Let $V_{max}$ and $\gamma$ be the sensor-specific design parameters where the maximum measurement that a sensor can register is represented by $V_{max}$ while $\gamma$ is a scaling factor corresponding to the sensor gain. The $t$th sample measurement of any sensor $n$ located at $(x_n, y_n)$ is given by
\begin{equation}\label{eq:1}
  z_{n,t} = min\{V_{max}, s_n + w_{n,t}\},
\end{equation}
for $n = 1, ..., N$, $t = 1, ..., M$, where attenuated signal $s_n$ is
\begin{equation}\label{eq:2}
  s_n = \gamma\frac{c}{r_n^\alpha},
\end{equation}
and $w_{n,t}$ is the additive white Gaussian noise, i.e., $w_{n,t} \sim \mathcal{N}(0, \sigma_w^2)$. In formula (\ref{eq:2}), the values of $\gamma$ and $\alpha$ are generally considered to be $1$ and $2$, respectively, and $r_n$ is the radial (Euclidean) distance from the source to sensor $n$, i.e.,
\begin{equation}\label{eq:3}
  r_n = \sqrt{(x_n - x_s)^2 + (y_n - y_s)^2}.
\end{equation}

Moreover, the $t$ represents the sample index while $n$ indexes the sensor nodes, and the samples are independent.

In a decentralized system, each sensor node generally quantizes the signal using preprogrammed threshold $T$ which is assumed to be common to all sensors in this paper. Given $T$ for each $t$ (sample index), we can define the following:
\begin{itemize}
  \item \textit{Alarmed Sensors}: Any sensor with $z_{n,t} \geq T$.
  \item \textit{Nonalarmed Sensors}: Any sensor with $z_{n,t} < T$.
\end{itemize}

The alarmed sensors communicate high bit (i.e., 1) to the fusion center while the nonalarmed sensors send nothing. The fusion center assume the 0 bit for nonalarmed sensors. The threshold $T$ is chosen such that the sensors get alarmed due to signal not noise.

Next we define the Region of Influence (ROI): the area around the source \cite{Michaelides}.

\textbf{Definition 1 (Region of Influence (ROI)):} Given the threshold $T$, Region of Influence (ROI) of a source represents the area around the source location inside which a sensor node gets alarmed with high probability, at least 0.5 (non-ideal situation).

\begin{rem}
The alarmed probability being at least 0.5 in definition 1 is due to symmetric PDF of noise \cite{Michaelides}. For uniform propagation model, ROI has the radius $R_c = \sqrt[\alpha]{\frac{\gamma c}{T}}$.
\end{rem}

\subsection{Fault Model}
We assume that the area, where the source is located, is densely populated (more enough) with sensor nodes. Each sensor node can exhibit erroneous behavior with probability $P_f$ which is bounded in the interval $[0, 0.5)$ generally, otherwise, sensor nodes will be replaced from the monitored area if their fault probabilities are greater.  Due to the fault model, sensor nodes' original decisions are simply reversed, leading to false positive and false negative decisions.

\begin{figure}[t]
\centering
\includegraphics[width=2.5in]{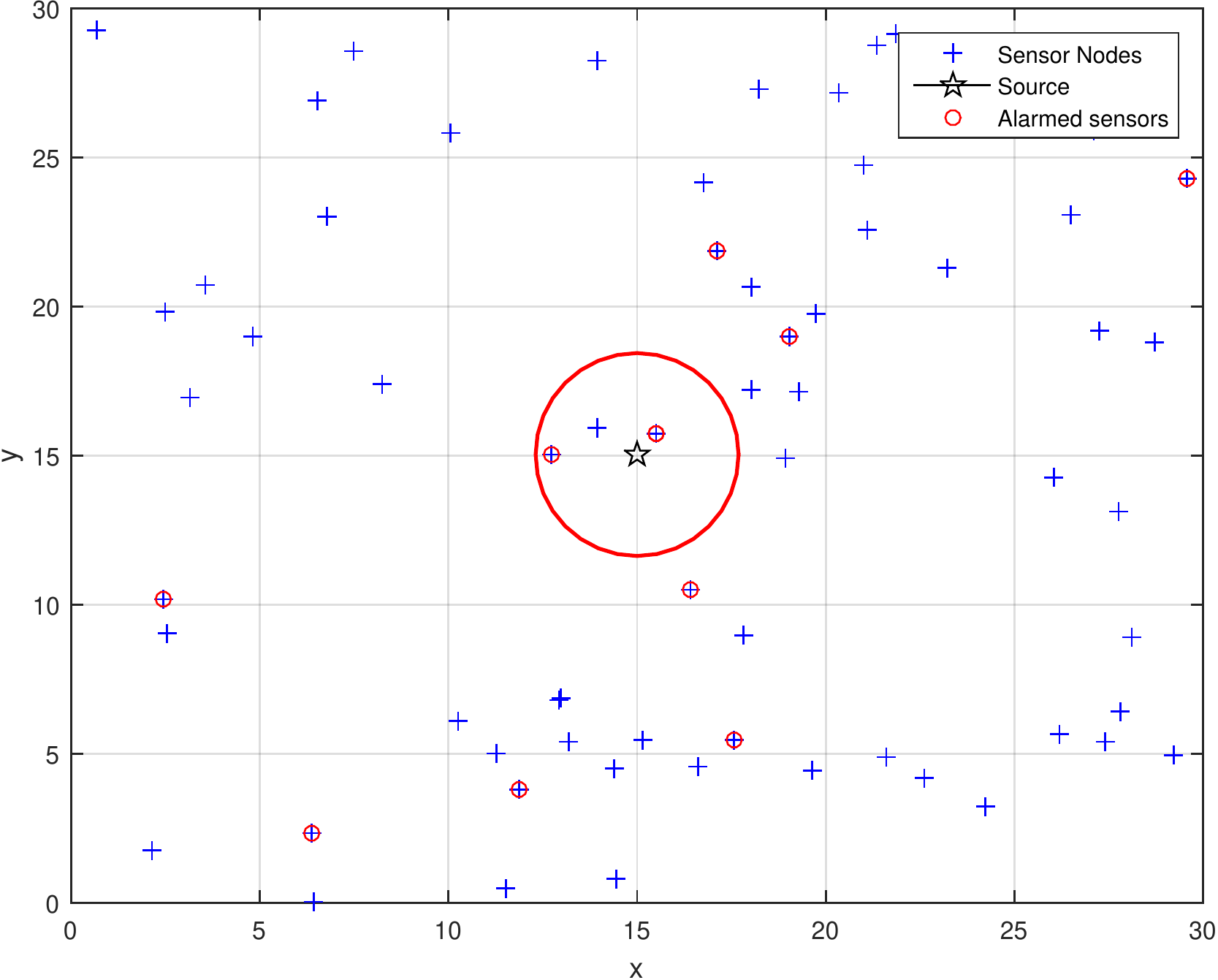}
\caption{An area covered with 65 uniformly random deployed sensor nodes. Source $s$ is place at (15, 15) represented by the star, and the big red circle around the source represents its ROI region.}
\label{source}
\end{figure}

Some sensors, located outside of ROI region, must not get alarmed, however, they get alarmed (false positive) while some sensors do not get alarmed despite living inside ROI region of the source but must get alarmed (false negative) as shown in Fig. \ref{source}. We will discuss in the problem statement that this fault model is equivalent to binary symmetric channels between sensors and the fusion center with probability $P_f$ as shown in Fig. \ref{model}.

\subsection{Related Work for Source Localization and their Disadvantages}
Although, many works have been done to estimate the source (event) location, following energy based estimators are the most suitable in the sense that they have low computation and communication complexities and they are \textit{\textbf{binary estimators}}. Moreover, they have been proposed or referred in \cite{MP}-\cite{Xu}.

\subsubsection{Centroid Estimator (CE)} This is the most basic estimator that computes the arithmetic mean of alarmed sensor nodes's coordinates.
Assume $(x_n, y_n)$, $n = 1, ..., P (P \leq N)$, denotes the positions of alarmed sensor nodes, then estimated source location is the centroid of these positions as given by
\begin{equation}\label{eq:4}
  \hat{\theta}_{CE} = [\hat{x}_s, \hat{y}_s] = \left[\frac{1}{P}\sum_{n=1}^{P} x_{n}, \frac{1}{P}\sum_{n=1}^{P} y_{n} \right ].
\end{equation}

Centroid estimator (\ref{eq:4}) assumes that each sensor's position is  equally important, hence equal weight, which makes it sensitive to false positive.
These faults make the estimated location $\hat{\theta}_{CE}$ to get sometimes far away from the actual source location. Generally, the centroid $\hat{\theta}_{CE}$ is calculated with only one sample, e.g., $M = 1$. For $M > 1$, first we calculate $\hat{\theta}_{CE}$ for each sample, then we average the Euclidean distances between the estimated location for each sample and the true source location.

\subsubsection{Maximum Likelihood (ML)} This estimator is defined in \cite{MP}-\cite{Xu} where it has been shown that this estimator is very
sensitive to false negatives. Moreover, only a single false negative can diverge the estimation result. The log likelihood function is given by
\begin{equation}\label{eq:5}
\begin{aligned}
  L(\theta) & = \sum_{n = 1}^{N}\sum_{t = 1}^{M}\bigg[I_{n,t} \log \left(Q(\frac{T - s_n(\theta)}{\sigma_w})\right)\\ & + (1 - I_{n,t})\log \left(1 - Q(\frac{T - s_n(\theta)}{\sigma_w})\right)\bigg],
\end{aligned}
\end{equation}
where $I_{n, t}$, $T$, and $\sigma_w$ represent the binary decision of sensor $n$ for sample $t$, threshold, and standard deviation of measurement noise $w_{n,t}$, respectively. Moreover, $Q(.)$ is the complementary distribution function of the standard Gaussian distribution and $s_n(\theta)$ is the measured signal of sensor $n$ without noise when the source is at location $\theta$.

\subsubsection{Fault Tolerant Maximum Likelihood (FTML)} FTML is proposed in \cite{MP} by taking into account the fault probability $P_f$ of each sensor,
and it reduces the effect of faulty sensors (false negatives) on the maximum likelihood estimator in formula (\ref{eq:5}). FTML requires the knowledge of $P_f$ (or its estimation), however, the fault probability $P_f$ is unknown in general and each sensor node may have different fault probability especially in heterogeneous sensor network (e.g., IoT networks \cite{Kumar}). To estimate the fault probability for each sensor node in heterogeneous sensor network is tedious task, so, the usage of FTML should be avoided in that situation.

In addition, it can be noted that after some period of operation of a sensor network for source localization, some sensor nodes can have different fault probability due to the power loss issues. Specifically sensor nodes which are closer to the source will work much of the time because they detect the source mostly and communicate their decisions to the fusion center. Hence they have more probability to become faulty comparatively.

\subsubsection{Subtract on Negative Add on Positive (SNAP)} This algorithm is proposed in \cite{Michaelides} by bounding the contribution of each sensor to be $\pm 1$. This algorithm uses only the sensor nodes' decisions which are in the ROI of the source, hence, false alarmed sensors will have no contribution. SNAP algorithm has two hurdles: the first is the assumption of knowledge (or estimation) of the ROI of the source and the second is its performance which is highly dependent on the grid resolution (technical parameter). Increasing grid resolution makes the algorithm less complex in the sense that it has small likelihood matrix  but it also deteriorates its performance.

Similarly, after some period of operation of a sensor network for source localization, the ROI can change (get larger) because of power loss issues of closer sensor nodes to the source. Closer sensor nodes will become faulty mostly, and hence, ROI gets larger which in turn degrades the performance of SNAP algorithm.

\subsection{Problem Statement} 
\begin{figure}[t]
\centering
\includegraphics[width=3in]{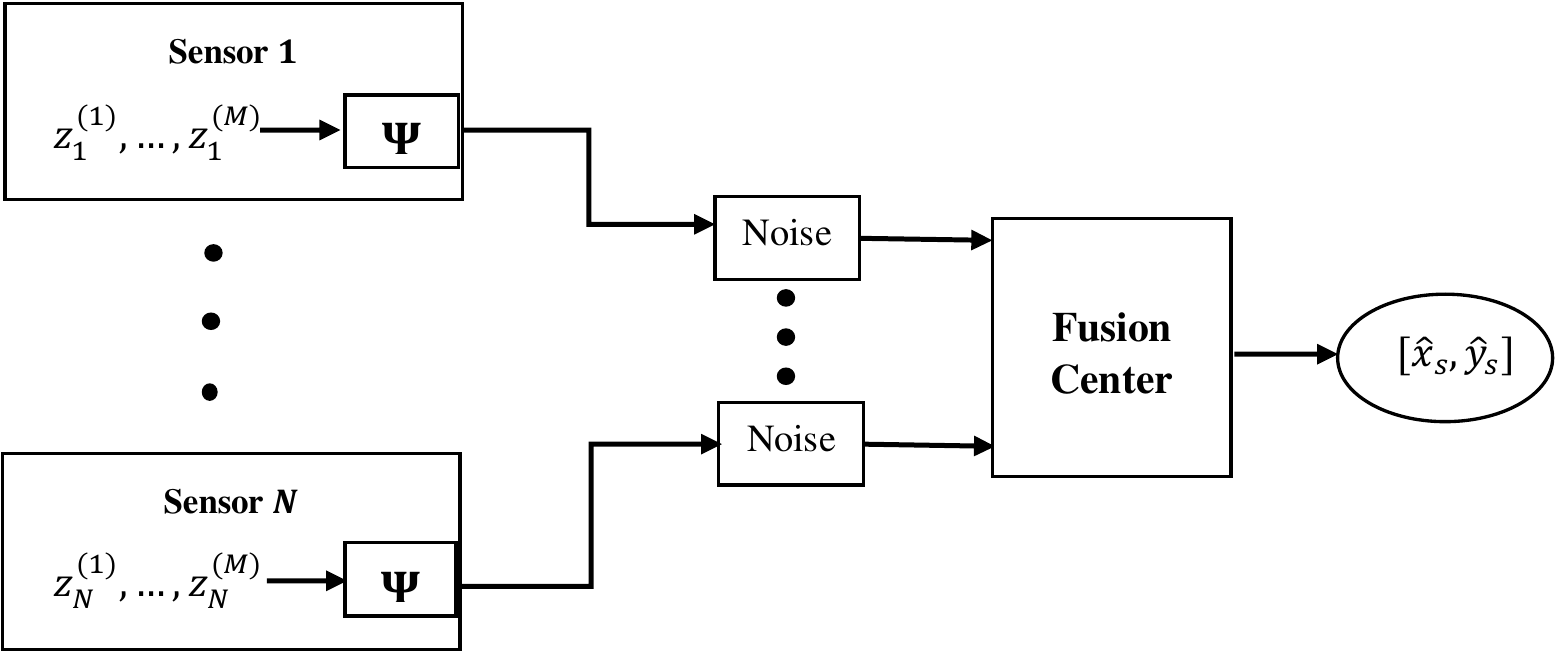}
\caption{Decentralized Source Localization Model.}
\label{model}
\end{figure}

Fig. \ref{model} shows the decentralized source location estimation model where sensor nodes quantize (i.e., quantizer $\Psi$) the measurements before transmitting them to the fusion center. The objective of this paper is to estimate the location of the source (event) under the fault model (noisy data set). Moreover, the sensor nodes quantize the real measurement into binary bit as we have discussed in Observation Model section. During the transmission of these bits to the fusion center, they go through the unreliable channels. We assume the binary symmetric channels with probability $P_f$, and these binary symmetric channels correspond to fault model of sensors where sensor nodes's decisions reverse with probability $P_f$ (e.g., false positives and false negatives).

We have discussed in Section 2.4 that the centroid estimator (CE) and maximum likelihood estimator are sensitive to faults (false positive and false negative, respectively). FTML takes into account the probability of each sensor node to be faulty, which is generally unknown and fault probability may be different for each sensor node especially in heterogeneous sensor networks. Although, the SNAP is a simple algorithm, it assumes the knowledge (or estimation) of the ROI of the source. Moreover, its performance is highly dependent on the gird resolution (technical parameter) as mentioned previously.

Hence, our objective is to estimate the source location using only the noisy data set when we do not have the knowledge (or estimation) of ROI of the source and fault probability $P_f$. Therefore, we consider all the alarmed sensor nodes' contributions independent of their positions (they may be anywhere). Furthermore, it can be shown that if we estimate the Region of Influence (ROI) of the source accurately, then it is equivalent to source location estimation because after estimation of ROI, we can use the centroid estimator (CE) to find the source location. Therefore, we will estimate the ROI of the source given the noisy data set.

\subsection{Hitting Set}
In this subsection, we define the hitting set, minimum hitting set, and minimal hitting set. These definitions can be found in \cite{Li}, \cite{Yang}.

\textbf{Definition 2 ((Minimum) Hitting Set):} Given a collection $\mathcal{C}$ of  some subsets of a set $P$, hitting set $H$ is a set that intersects each subset in the collection with at least one element. A \textit{minimum hitting set} is a hitting set of the smallest size.

\noindent \textbf{Example 1 (Minimum Hitting Set):} We have a collection $\mathcal{C} = \{\{1\}, \{1, 4\}, \{2, 3\}, \{2, 4\}\}$ of subsets. The minimum hitting set of this collection is $\{1, 2\}$.

\textbf{Definition 3 (Minimal Hitting Set):} A hitting set of a collection of some subsets of a set is \textit{minimal} if and only if no proper subset of it is a hitting set for this collection.

\noindent \textbf{Example 2 (Minimal Hitting Set):} Consider the collection $\mathcal{C} = \{\{1, 2\}, \{1, 3\}, \{1, 2, 4\}, \{1, 3, 5\}\}$. Using definition 3, one of the minimal hitting sets of this collection is $\{2, 3\}$ because no proper subset of this is a hitting set.

\begin{rem}
  It can be proved that the minimum hitting set is also the minimal hitting set but not vice versa \cite{Li}. For instance, in Example 2, there exists the minimum hitting set $\{1\}$ which is also the minimal hitting set.
\end{rem}

\section{Source Localization under the Fault Model}
In this section, we propose two methods to recover the ROI of the source, hence, the source location. We show that the proposed methods (hitting set approach and feature selection method) can recover the ROI of the source, and these methods are more fault tolerant as compared to CE and ML. We also derive the lower bound on the sample complexity (Proposition 1) for hitting set approach.

Let $\mathcal{D} = \{\textbf{x}^1, ..., \textbf{x}^M\}$ be the data set received at the fusion center from all the sensor nodes through noisy channels where $\textbf{x}^i \in \{0, 1\}^N$ and superscript $i$ is the time index, and $M$ represents the total number of samples.

Under the fault model where each sensor's decision is reversed with probability $P_f$, our goal is to estimate the source location using the noisy data set (OR estimate the Region of Influence (ROI) based on source neighborhood). We also assume that the environment is densely populated (more enough) with sensor nodes. Multiple sensors observe the source, consequently, spatial proximal sensor observations are highly correlated \cite{Liu}-\cite{Krishna}.

Let $G(V, E)$ be a graph with vertex set $V = \{1, ..., N+1\}$ and edge set $E = \{(i, j) : \forall i, j \in V\}$. Vertex set V represents the sensor nodes, and we also consider the source $s$ as a node in the graph. There exists an edge between the source and a sensor node if the sensor node gets alarmed, e.g., $N(s) = \{i : (s, i) \in E\}$ representing the neighborhood of the source $s$. Due to high density (more enough) of sensors in the area of the source, the sensors are dependent showing the spatial correlation, and they become conditionally independent given the source observation. Our goal is to find the true source's neighborhood $N(s)$ given the noisy binary data set.

First we show that the ROI of the source is equivalent to its neighborhood in Lemma 1. We assume the ideal situation such that if a sensor node is inside the ROI of the source, it will get alarmed definitely without any noise and fault model.

\textbf{Lemma 1 (Equivalent Definition of ROI):} Suppose no noise (e.g., $w_{n, t} = 0$) and no fault ($P_f = 0$), assuming 1) a uniform propagation model (as in formula (2)) and 2) the presence of at least one sensor node inside the source ROI, the source position can be estimated based on the decisions of sensors inside its ROI which is equivalent to $ROI \equiv N(s)$ where $N(s)$ is the set of sensors (true neighborhood of the source $s$) that get alarmed.

\begin{rem}
  In Lemma 1, assumption 1 is due to the model where source can be detected by sensors around it while assumption 2 is to exclude the chance of undetected source or there is no sensor nearby to the source.
\end{rem}

\begin{proof}
  Since Region of Influence (ROI) is explained in Definition 1. Under the conditions of no noise and no fault, the sensors will detect the source if these sensors are inside the ROI of the source $s$. Source location can be estimated by centroid of the decisions of sensors around its ROI.

  Moreover, let $r$ be the radius of the ROI region. By uniform propagation model assumption, $ROI \equiv B_r(s) = \{i : d(s, i)\leq r\}$ is the set of sensors which get alarmed where $d(s,i)$ is the Euclidean distance. We model the source $s$ as a node in the graph $G(V, E)$, and the set of alarmed sensor nodes inside the ROI, which have detected the source, is denoted by $N(s) = \{i : (s, i) \in E\}$. We need to show that this neighborhood is equal to $B_r(s)$.

  We try to prove that $N(s) \subseteq B_r(s)$ by contradiction method: we know that if $j \in N(s)$, then $j \in B_r(s)$. Assume  $j \not \in B_r(s)$, then $d(s, j) > r$ which means sensor node $j$ is located outside the ROI of the source. Then $(s, j)$ is not an edge in the set $E$ because $j$ will not get alarmed if it is outside the ROI of the source, therefore, $j \not \in N(s)$. Hence, $N(s) \subseteq B_r(s)$.

  Similarly, we can prove that $B_r(s) \subseteq N(s)$. Therefore, $N(s) = B_r(s)$.
\end{proof}

Hence, to estimate accurately the source position implies the accurate estimation of source neighborhood under the fault model given the noisy data set $\mathcal{D}$.

\subsection{Hitting Set Approach}
Given a sample $\textbf{x}$, $\hat{N}(s)$ represents the estimated neighborhood of the source $s$, which is defined as $\hat{N}(s) = \{i: x_i = 1\}$ where $x_i = 1$ is the $ith$ sensor's decision in the sample $\textbf{x}$. Since there are $M$ samples in the data set, there will be an estimated collection of neighborhoods for the data set, e.g., $\{\hat{N}^{(k)}(s): k =1, 2, ..., M\}$. Hitting set approach has been used previously in \cite{Li}, \cite{Yang} for epidemic graph learning and interference graph learning of a wireless network, respectively. However, we employ hitting set approach for source localization in this paper.

We assume that each sample $\textbf{x}$ contains at least one correctly alarmed sensor node (or correct decision of a sensor node), which is the reasonable assumption due to the fact that the fault probability $P_f$ is generally bounded (i.e., $P_f \in [0, 0.5)$) and dense deployment (more enough) of sensor nodes. Theorem 1 proves that $N(s)$ (true source neighborhood) can be recovered using the minimum hitting set approach.

\textbf{Theorem 1:} Consider the data set $\mathcal{D}$ received at the Fusion Center through BSC channels with fault probability $P_f \in [0, 0.5)$ as shown in Fig. \ref{model}. Assume that each sample $\textbf{x}$ contains at least one correctly alarmed sensor decision, then $N(s)$ (i.e., the true source neighborhood) is (equal to) the unique minimum hitting set given the collection of estimated neighborhoods $\{\hat{N}^{(k)}(s)\}$ as $k \rightarrow \infty$, where $k$ represents the sample number.
\begin{proof}
  First we prove that the $N(s)$ is the hitting set and then prove that it is the unique minimum hitting set.
  \begin{enumerate}
    \item Since each sample contains at least one correctly alarmed sensor decision, $N(s) \cap \hat{N}^{(k)}(s) \not = \phi$ for $\forall k$. Hence $N(s)$ is the hitting set of the collection $\{\hat{N}^{(k)}(s)\}$.
    \item Suppose that $N(s)$ is not the minimum hitting set, then there exists at least one different hitting set $\tilde{N}(s)$ whose size is smaller than
    $N(s)$, e.g., $|\tilde{N}(s)| \leq |N(s)|$. So there must exist at least one sensor node $l \in N(s)$ but $l \not \in \tilde{N}(s)$.

    Consider an event such that there exists at least one estimated neighborhood of the source $\hat{N}^{(k)}(s)$ given the data set, which contains only one sensor node $l$ (e.g., $l \in \hat{N}^{(k)}(s)$).
    Although, this event has a small probability, it will definitely occur as $k \rightarrow \infty$ because some sensor nodes are located at the boundary of the ROI denoted as a circle. Suppose $H$ sensor nodes including sensor node $l$ are located at the boundary of the ROI, then $l \in \hat{N}^{(k)}(s)$ is possible due to the assumption that at least one correctly alarmed sensor node exists in each sample and the fault probability of sensor nodes $P_f$ which can reverse the decisions of remaining sensor nodes in $H$ but could not reverse the decisions of nonalarmed sensor nodes. When this event occurs in any sample, suppose sample k, then $\tilde{N}(s) \cap \hat{N}^{(k)}(s) = \phi$, which contradicts that $\tilde{N}(s)$ is the hitting set.
  \end{enumerate}

  Therefore, we conclude that $N(s)$ is the minimum hitting set as $k \rightarrow \infty$.
\end{proof}

\begin{rem}
  Similarly, we can use the minimal hitting set to recover $N(s)$ \cite{Li} because it is known that finding the minimum hitting set of a collection of some subsets is NP-complete, however, we can find a minimal hitting set of the collection in polynomial time.
\end{rem}

We have proved in Theorem 1 that $N(s)$ can be recovered as $k \rightarrow \infty$ using the minimum hitting set. Following proposition gives the lower bound on the number of samples required to learn $N(s)$.

Let $\hat{N}^{(M)}(s)$ denote the estimated minimum hitting set (estimated neighborhood of the source $s$) having observed $M$ samples, and $E$ denotes an event such that the estimated minimum hitting set (from $M$ samples) is not equal to true source neighborhood $N(s)$. It only occurs when $|\hat{N}^{(M)}(s)| \leq |N(s)|$ using Theorem 1, otherwise $\hat{N}^{(M)}(s)$ is the feasible estimated minimum hitting set in the sense that it recovers the neighborhood of the source $s$ with finite sample size.

\textbf{Proposition 1 (Sample Complexity Bound):} Let $\delta > 0$, and let M be the number of samples which should be
\begin{equation}\label{eq:6}
  M \geq \frac{\log \delta - \log d}{\log P_f},
\end{equation}
then with probability at least $1 - \delta$, $\hat{N}^{(M)}(s) = N(s)$ given binary data set with $M$ samples, where $\hat{N}^{(M)}(s)$ and $N(s)$ denote the estimated minimum hitting set and true source neighborhood, respectively. Moreover, $d$ represents the degree of $N(s)$ and $P_f$ is the fault probability of each sensor node. This fault probability can be considered as the BSC channels error probability, which exist between sensors and the fusion center as shown in Fig. \ref{model}.

\begin{proof}
When $|\hat{N}^{(M)}(s)| \leq |N(s)|$, there may exist sensor node $l \in N(s)\backslash \hat{N}^{(M)}(s)$, therefore, the probability of error of all such sensor node $l$ is as follows.
\begin{equation}\label{eq:7}
\begin{split}
    Pr(\hat{N}^{(M)}(s) \not= N(s)) & = Pr(\cup _{l \in N(s)} l \not \in \hat{N}^{(M)}(s)), \\
    & \stackrel{(a)}{\leq} \sum_{l \in N(s)} Pr(l \not \in \hat{N}^{(M)}(s)), \\
    & \stackrel{(b)}{\leq} d \cdot (P_f)^M,
\end{split}
\end{equation}
where $(a)$ and $(b)$ are due to union bound and the fault probability of each sensor $P_f$ for every sample and the degree of the source $|N(s)| = d$, respectively.  The fault probability $P_f$ makes the sensor decision to reverse. The reversal of the sensor decision may cause the sensor node $l$ decision not to be included in the estimated minimum hitting set. Then $Pr(l \not \in \hat{N}^{(M)}(s)) \leq (P_f) ^M$ for $M$ number of samples. Taking logarithm of formula (\ref{eq:7}-(b)) will give the required result.
\end{proof}

Having estimated the ROI of the source $s$, the estimated source location will be the centroid of the sensor nodes in the ROI as mentioned in the problem statement. Let $|\hat{N}^{(M)}(s)| = d$ be the number of sensor nodes in the estimated ROI (neighborhood of the source (Lemma 1)), then the estimated source location is given by
\begin{equation}\label{eq:8}
  \hat{\theta}_{HS} = [\hat{x}_s, \hat{y}_s] = \left[\frac{1}{d}\sum_{n=1}^{d} x_{n}, \frac{1}{d}\sum_{n=1}^{d} y_{n} \right],
\end{equation}
where $(x_n, y_n)$ is the true location of the $nth$ sensor node.

\begin{figure}[t]
\centering
\includegraphics[width=2.5in]{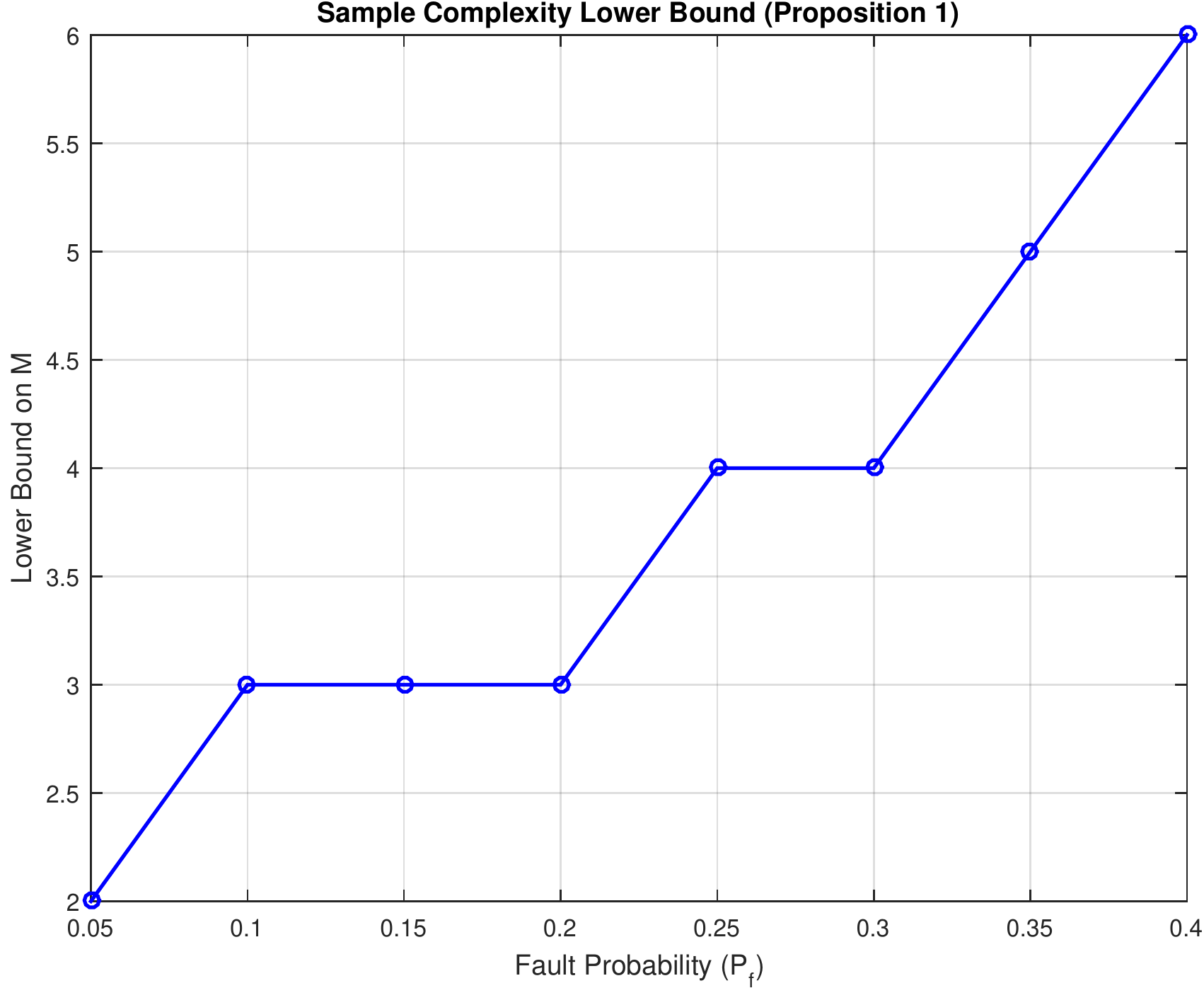}
\caption{Sample complexity requirement for Example 3.}
\label{example}
\end{figure}

\noindent \textbf{Example 3 (Sample Complexity Lower Bound):} In this example, we show the sample requirement (Proposition 1) for different values of sensor nodes fault probability when the source $s$ has $d = 10$ neighboring sensor nodes, and $\delta = 0.1$. Moreover,  the condition $d = 10$ is reasonable for densely populated monitored area. As shown in Fig. \ref{example}, the sample complexity increases with the fault probability of sensor nodes. This sample requirement is necessary to achieve $90\%$ success rate but it is not sufficient requirement.

\subsection{Feature Selection Approach}
Given the noisy data set $\mathcal{D} = \{\textbf{x}^1, ..., \textbf{x}^M\}$, our goal is to estimate the location of the source using wireless sensor network. The noisy data set is generated due to fault model with probability $P_f$. In this subsection, we propose an algorithm to estimate the neighborhood of the source $s$ using the noisy data set, then localize the source using the estimated neighborhood.

Algorithm 1 has three main components: counting the frequency of each alarmed sensor node in the data set, selecting the most relevant feature (sensors), and estimating the source location.

\subsubsection{Counting Function} Given the data set $\mathcal{D}$ with $M$ samples, we find out the frequency of alertness of each sensor (line 14-22), and save and return the output.

\subsubsection{Feature Selection}
Having found out the frequency of each sensor alertness, we select the most relevant features which are those sensor nodes which get alarmed frequently. Since the fault model probability $P_f$ is bounded generally in the interval $[0, 0.5)$, sensor nodes, which are in the ROI of the source, get alarmed most of the time.

Our feature selection criteria (line 3-11) is to exclude those sensors which did not get alarmed $25\%$ of the number of samples. This criteria is chosen to reduce the effect of false alarmed sensors on the source location. Moreover, other criteria may be used.

\subsubsection{Source Location Estimation} After selection of relevant sensors (e.g., $B, C, D$ arrays), we need to estimate the source location given the weights, i.e., $W$. The weight sequence $W$ is user-defined to account for the fact that frequently alarmed sensor nodes will get higher weights in centroid calculation. Source location is estimated by the weighted centroid of the relevant sensors (line 23-30) because these arrays (e.g., $B, C, D$) contain sensors with different degrees of relevance, hence, we need to take into account their relevance levels to the source location. This will further reduce the impact of false alarmed sensors.

Although, this method has some similarities with feature selection method, which is used to reduce the dimensionality of the data set in different learning problems such as \cite{Bennasar}, it is different in the sense that we select features (sensors) based on the frequently alarmed sensors given the number of observations while feature selection methods in the literature consider the relevance of features with the class label.

\subsubsection{Analysis of Algorithm 1}
\textbf{Time Complexity:} The overall time complexity of Algorithm 1 is $\mathcal{O}(MN)$ where $M$ and $N$ represent the number of samples and number of sensors, respectively. In addition, Count function dominates this time complexity. Moreover, the time complexity of Centroid function is logarithmic (Line 23-30) while the complexity of Line 3-11 is $\mathcal{O}(N\log N)$.

\noindent \textbf{Feature (Sensor) Selection Criteria (line 3-11):} We select those sensor nodes which get alarmed mostly because of the fact that the sensors which are in ROI of the source get alarmed frequently despite the fault model. Having selected those sensor nodes, we assign the highest weight to them for source location estimation.
For instance, consider Centroid Function where after selecting the most relevant features (i.e., $B, C, D$), we calculate the centroid given the input weight vector $W$. The weight vector is user-defined sequence of weights to account for the fact that frequently alarmed sensor nodes will get higher weights in centroid calculation.  Having this criteria will diminish the effect of false alarmed sensors on the centroid (source location).
\begin{algorithm}[hbt!]
\LinesNumbered
\SetAlgoLined
\KwIn{Data set: $\mathcal{D} = \{\textbf{x}^1, ..., \textbf{x}^M\}$, One dimensional weight array: $W$}
\KwOut{Estimated source location: [$\hat{x}_s, \hat{y}_s$]}
{
Initialization: One dimensional array $A$ for all sensors' decisions, and one dimensional $B, C, D$ arrays for storing relevant sensor nodes\;
A = Count($\mathcal{D}$) \tcp*[r]{counting the alarmed sensors' decisions}
\For{$n = 1$ to $N$}{
 \uIf{$A[n] \geq \lfloor\frac{3M}{4}\rfloor$}{
    B.append($n$) \tcp*[r]{add on the sensor $n$ in array B}
    }
    \uElseIf{$\lfloor\frac{M}{2}\rfloor \leq A[n] < \lfloor\frac{3M}{4}\rfloor$}{C.append($n$) \tcp*[r]{add on the sensor $n$ in array C}}
    \uElseIf{$\lfloor\frac{M}{4}\rfloor \leq A[n] < \lfloor\frac{M}{2}\rfloor$}{D.append($n$)\tcp*[r]{add on the sensor $n$ in array D}}
    \textbf{end}
}
$[\hat{x}_s, \hat{y}_s] = $ Centroid (B, C, D, W) \tcp*[r]{estimated location of the source $s$}
}
\textbf{Functions:}

\SetKwFunction{FC}{Count}
\SetKwFunction{FCD}{Centroid}
\SetKwProg{Fn}{Function}{:}{}
    \Fn{\FC{$\mathcal{D}$}}{
    sum[] = 0 \tcp*[r]{array for all sensors, $M:$ total number of samples, $N:$ total number of sensors}
    \For{$k = 1$ to $M$}{
    \For{$n = 1$ to $N$}{
    sum[$n$] += $x[k][n]$ \tcp*[r]{sum all the decision of each sensor for M samples}
    }
    }
    \Return sum}
    \textbf{End} Count

    \Fn{\FCD{$B, C, D, W$}}{
    $P = |B|, Q' = |C|, R = |D|$ \tcp*[r]{let the sizes of the arrays be, and array W contains the weight for each array}

    $[\hat{x}_1, \hat{y}_1] = \left [\frac{1}{P}\sum_{g = 1}^{P}x_g, \frac{1}{P}\sum_{g = 1}^{P}y_g \right ]$ \tcp*[r]{$g \in B$ and each sensor corresponds to its position. $[\hat{x}_l, \hat{y}_l]$ is the estimated source location for any $l$}

    $[\hat{x}_2, \hat{y}_2] = \left [\frac{1}{Q'}\sum_{h = 1}^{Q'}x_h, \frac{1}{Q'}\sum_{h = 1}^{Q'}y_h \right ]$ \tcp*[r]{$h \in C$}

    $[\hat{x}_3, \hat{y}_3] = \left [\frac{1}{R}\sum_{i = 1}^{R}x_i, \frac{1}{R}\sum_{i = 1}^{R}y_i \right ]$ \tcp*[r]{$i \in D$}

    $[\hat{x}_s, \hat{y}_s] = \left [\frac{1}{3}\sum_{l = 1}^{3}w_l \hat{x}_l, \frac{1}{3}\sum_{l = 1}^{3}w_l \hat{y}_l \right ]$ \tcp*[r]{final weighted average estimated location where $w_l \in W$}
    \Return $[\hat{x}_s, \hat{y}_s]$
    }
    \textbf{End} Centroid
\caption{Feature Selection Method for Source Localization}
\label{Algorithm}
\end{algorithm}

Next we prove that Algorithm 1 will reduce the effect of false alarmed sensor nodes on the estimated source location.

\textbf{Theorem 2:} Algorithm 1 gives the feasible solution in the sense that it returns the source location which is in the monitored area A. Moreover, it reduces the effect of false alarmed sensors (false positives) on the source location in comparison to Centroid Estimator (CE) without feature selection.
\begin{proof}
  We know that the sensor nodes in the ROI of the source get alarmed frequently despite having the fault model with probability $P_f$. Due to the fault probability $P_f$, the decisions of sensors reverse, e.g., false negatives and false positives.

  The feasibility of Algorithm 1 can be easily shown. For instance, it selects the frequently alarmed sensors (line 2-22), and then estimates the source location using the selected sensor nodes (line 23-30). The estimated location, in fact, is in the monitored area A because all the sensor nodes are located in the area.

  Algorithm 1 selects the three sets of features (line 3-11) depending on the frequency of alertness of sensors given the data set $\mathcal{D}$. Using this part of the algorithm (line 3-11), we exclude those sensors which did not get alarmed $25\%$ of the number of samples in the data set. This will reduce the effect of false alarmed sensors (e.g., fault model) on the estimated source location because those sensors get alarmed frequently which are in the ROI region of the source.

  Moreover, Centroid Function (line 23-30) also reduces the impact of fault model further because it calculates the weighted average of the selected features where more frequently alarmed sensors get more weights comparatively.
\end{proof}

\section{Extension to Multiple Sources Localization}
In addition of a single source localization problem, multiple sources localization problem has been studied as well in \cite{Laoudias}, \cite{Xu}-\cite{Xin}. In the literature, multiple sources localization is studied for same type of sources in general, where the measurements of multiple sources superimpose for a sensor node. Moreover, the shape and size of ROIs of these sources are dependent on the distance between the sources. For instance, the ROIs of two sources get connected if and only if the distance $d$ between them is $d \leq L$ \cite{Laoudias}, and $L$ is defined as
\begin{equation}\label{eq:9}
  L = \frac{1}{\sqrt[\alpha]{\gamma}} \left(\sqrt[\alpha + 1]{c_1} + \sqrt[\alpha + 1]{c_2}\right)^{\frac{\alpha + 1}{\alpha}},
\end{equation}
where $c_1, c_2$ are the emitted signals by the two sources, and $\alpha$ and $\gamma$ are design specific parameters that depend on the environment.

Furthermore, it has been shown that distributed SNAP localization algorithm fails to distinguish any two sources with distance $d \leq 2R_c$ in two dimension setting (in dense WSN, and $R_c$ is defined in Remark 1) \cite{Laoudias}.

In this section, we discuss the extensions of our proposed methods (hitting set approach and feature selection method) to localize $K$ sources.

In contrast to literature, we do not need to bound the distance $d$ between the sources, however, initial guess of clusters of sensor nodes around the sources is required. For instance, we can use mean shift hierarchical method \cite{Yu} or K-means clustering algorithm \cite{Sasikumar} to estimate the clusters of sensor nodes around the sources because the sensor nodes locations are known. Moreover, these clusters may not be disjoint due to connected ROIs of the sources. Our proposed extended methods in this section can also be used to localize unknown number of sources \cite{Xin}, though, the performances will be highly dependent on the clustering algorithm.

\subsection{Extension of Hitting Set Approach}
In Theorem 1, we have shown that the source neighborhood can be recovered if the number of samples goes to infinite. Let $\hat{C}(s_1), ..., \hat{C}(s_K)$ be the initial estimated clusters (using K-means clustering) of the $K$ sources where $\hat{C}(s_k) = \{sensor\; nodes\}$ for any $k$. These clusters are estimated once (at the start of the algorithm, and these clusters may not be disjoint as mentioned above), then hitting set approach is used for each source separately on the data related to sensor nodes in the estimated source cluster. For instance, the hitting set approach estimates the neighborhood of a source $s_1$ by only considering those sensor nodes (therefore their data) which are in the estimated cluster of the source $s_1$, e.g., $\hat{C}(s_1)$. The working condition of this approach is based on the dense (enough nodes) wireless sensor networks and the conditions assumed in Theorem 1.

For a single source localization problem, hitting set approach estimates the neighborhood of the source, on the other hand, this approach (extended version of the hitting set) estimates the structure of the graph (structure learning problem especially for connected ROIs of sources) for multiple sources localization where $N$ sensor nodes and $K$ sources are the nodes of the graph. We only estimate the neighborhoods of the sources because $K$ sources are conditionally independent given their neighborhoods.

\noindent \textbf{Sample Complexity Bound:} In addition, we can extend Proposition 1 (lower bound on the sample complexity) for $K$ sources as follows. We have discussed that we model the multiple sources localization problem as the structure learning problem,  and the sample complexity requirement is given by
\begin{equation}\label{eq:10}
  M \geq \frac{\log \delta - \log (Kd)}{\log P_f},
\end{equation}
where $\delta > 0$, and with probability at least $1 - \delta$, $\hat{G} = G$ given binary data set with $M$ samples. Furthermore, $\hat{G}$ and $G$ denote the estimated graph and true graph, respectively, and $d$ represents the degree of each source and $P_f$ is the fault probability of each sensor node. This fault probability can be considered as the BSC channels error probability, which exist between sensors and the fusion center as shown in Fig. \ref{model}.

\subsection{Extension of Feature Selection Method}
We first need the estimation of clusters of sensors around the sources using K-means clustering algorithm (e.g., as in Section 4.1), then Algorithm 1 can be used for each source separately using the data of those sensor nodes which are in the estimated source cluster. We also have discussed that clusters may not be disjoint, therefore, some sensor nodes may belong to multiple sources' ROIs.

\subsection{Iterative Multiple Sources Localizations Algorithm}
In Section 4.1 and 4.2, we utilize the estimated clusters once, which are estimated by mean shift method \cite{Yu} or K-means clustering \cite{Sasikumar}. After clusters estimation, we employ our proposed methods: hitting set approach or feature selection method, to localize multiple sources. The performances of the proposed methods for multiple sources localizations are highly dependent on the clustering algorithms.

In this subsection, we propose an iterative algorithm similar to K-means clustering to overcome the performance dependence on the clustering algorithms. The proposed algorithm is given in Algorithm 2 which is based on proposed hitting set or feature selection approach. The number of clusters is same as the number of sources.

In Algorithm 2, we uniformly assign each sensor node to a cluster, then use hitting set method or feature selection method for each cluster to estimate the sources locations. In addition, these two steps are repeated based on optimized clusters and sources locations, respectively, until required performances are achieved or no change in performances happens further.

\begin{algorithm}[hbt!]
\LinesNumbered
\SetAlgoLined
\KwIn{Sensor positions: $(x_n, y_n)$ where $n \in [N] = \{1, 2, ..., N\}$}
\KwOut{Sources estimated locations: [$\hat{x}_j, \hat{y}_j$] for $j = 1, ..., K$}
{
\textbf{Initialization:} Define $K$ clusters $C_1, C_2, ..., C_K$ for each source.

\textbf{Uniform Assignment:} Uniformly pick sensor nodes from the set $[N]$ and put it into the cluster $C_j$ uniformly for any $j$ such that no cluster is empty.

\textbf{Sources Locations:} Utilize Hitting set approach or Feature selection approach to each cluster $C_j$ for $j$th source localization.

\textbf{Clusters Optimization:} Cluster each sensor node again from the set $[N]$ into the clusters $C_1, C_2, ..., C_K$ based on Euclidean distance between estimated sources locations in step 3 and sensor nodes.

\textbf{Sources Locations Optimization:} Perform sources localizations based on Hitting set/Feature selection method using optimized clusters.

\textbf{Repeat Steps 4 and 5:} Repeat steps 4 and 5 until required performances are achieved or no change in performances happens.
}
\caption{Iterative Multiple Sources Localizations}
\label{Algorithmtwo}
\end{algorithm}

\section{Simulation Results}
All the simulation results are obtained by implementing the programs in Matlab. In addition, we use integer linear programming for finding the minimum hitting set. To improve the time complexity, genetic algorithm can be utilized for finding the minimum hitting set as well \cite{Bojana}. Furthermore, we set weight array $W = [1, 1, 1]$ for feature selection method as required in the final step of source estimation in Algorithm 1. This weight array behaves like the optimization parameter whose configuration must be selected according to the condition that the frequently alarmed sensor nodes get higher weights in the centroid calculation in Algorithm 1. For simplicity in experiments, we choose the weight array $W$ all ones vector.

Following observation model is used for the simulations in the sensor field of $100 \times 100$ area A:
\begin{equation}\label{eq:11}
  z_{n, t} = min\{3000, \sum_{j = 1}^{K}s_{n, t}^j + w_{n, t}\},
\end{equation}
where the attenuated signal is $s_{n, t}^j = \frac{c^j}{r_{n, j}^2}$, and $w_{n, t}$ is the additive white Gaussian noise having zero mean and $\sigma_w^2$ variance. Moreover, $c^j = 3000$ is the emitted signal from source $j = 1, ..., K$ and $r_{n, j}$ is the Euclidean distance between sensor $n$ and source $j$ as defined in formula (\ref{eq:3}).

In addition, the average root mean square (RMS) error is used as the performance metric, which can be defined as
\begin{equation}\label{eq:12}
\begin{multlined}
  Average \: RMS \: Error = \\ \frac{1}{BK}\sum_{j = 1}^{K}\sum_{b = 1}^{B}\sqrt{(\hat{x}_{(s_j, b)} - x_{s_j})^2 + (\hat{y}_{(s_j, b)} - y_{s_j})^2},
\end{multlined}
\end{equation}
where $j$ and $B$ are the source index and the number of simulations performed, respectively. We set the value $100$ (e.g., $B = 100$) for the number of simulations performed for each experiment in this paper.

In Section 5.1, we simulate our results for a single source localization in comparison with Centroid Estimator (CE) and Maximum Likelihood (ML) estimator, and moreover, we also compare our results with FTML \cite{MP} and SNAP algorithm \cite{Michaelides} in Section 5.2 for a single source. Although, Fault Tolerant ML and SNAP algorithm  achieve the best known performances for source localization in \textbf{general case}, they take into account strong assumptions as pointed out in Section 2.4 in this paper, which are not easy for practical implementation. In Section 5.3, we simulate our results for two sources localization problem.

\subsection{Simulation for a Single Source}
In this subsection, we perform extensive simulations to validate our theoretical results. We compare our estimators with Centroid Estimator (CE) formula (\ref{eq:4}) and Maximum Likelihood (ML) estimator formula (\ref{eq:5}).

\subsubsection{Optimal threshold value}
In this subsection, we need to find the optimal threshold $T$ for a single source such that the performance is maximum under the fault model for different number of sensor nodes and sample complexity. Threshold $T = 5$ has been used previously for the observation model in formula (\ref{eq:11}) in \cite{MP}, \cite{Michaelides} where the threshold value is derived analytically and experimentally. However, we have only the noisy data set in this paper, so, the optimal value of threshold T is found experimentally. To find the optimal threshold $T$ under the measurement noise and the fault model, we consider the observation model in formula (\ref{eq:11}) with $w_{n, t} \sim \mathcal{N}(0, 1)$ and the fault probability $P_f = 0.1$ for each sensor node.

\begin{figure*}[ht]
\centering
\begin{subfigure}{.4\textwidth}
  \centering
  \includegraphics[width=2.5in]{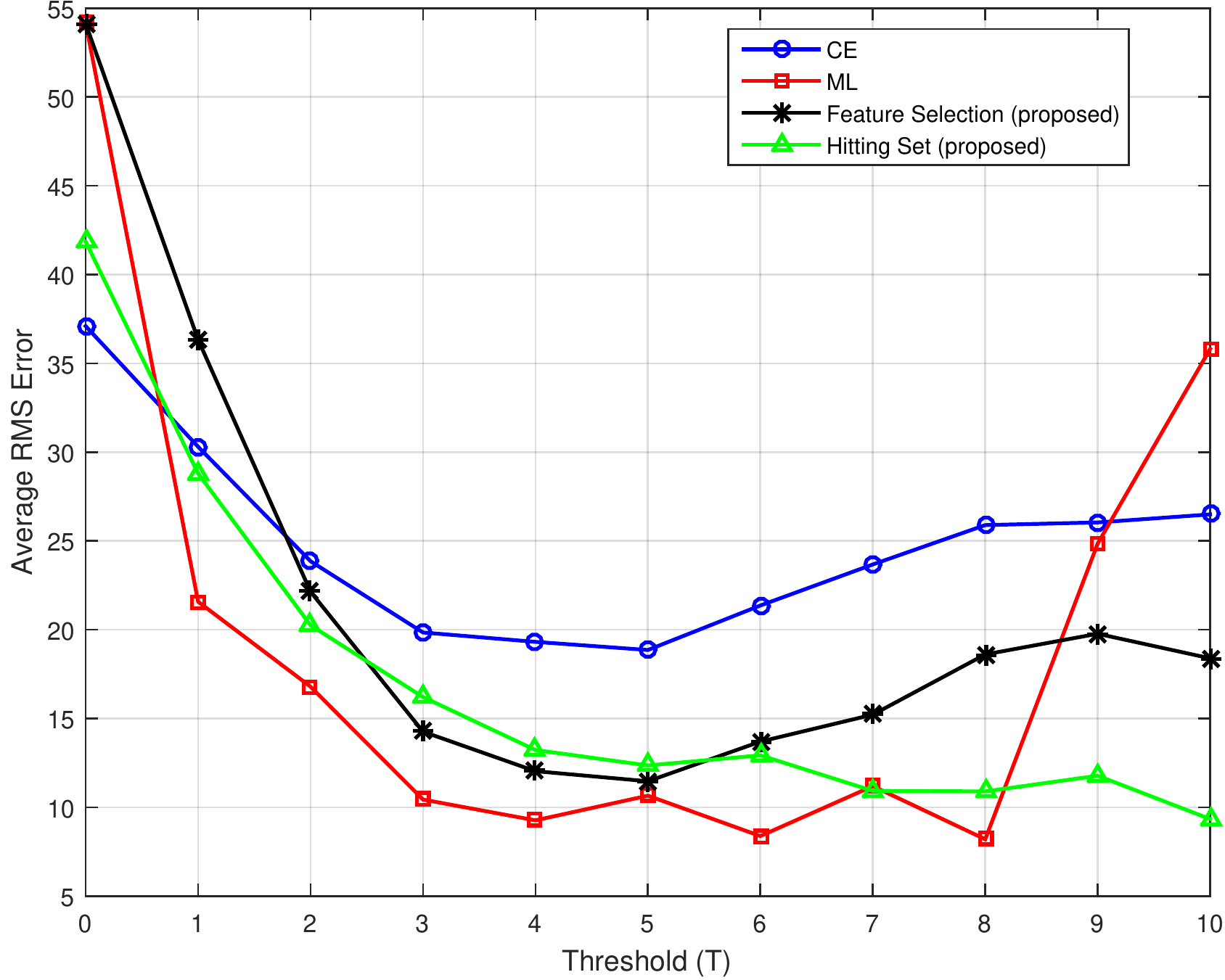}
  \caption{$M = 50, N = 150$}
  \label{n150}
\end{subfigure}
\begin{subfigure}{.4\textwidth}
  \centering
  \includegraphics[width=2.5in]{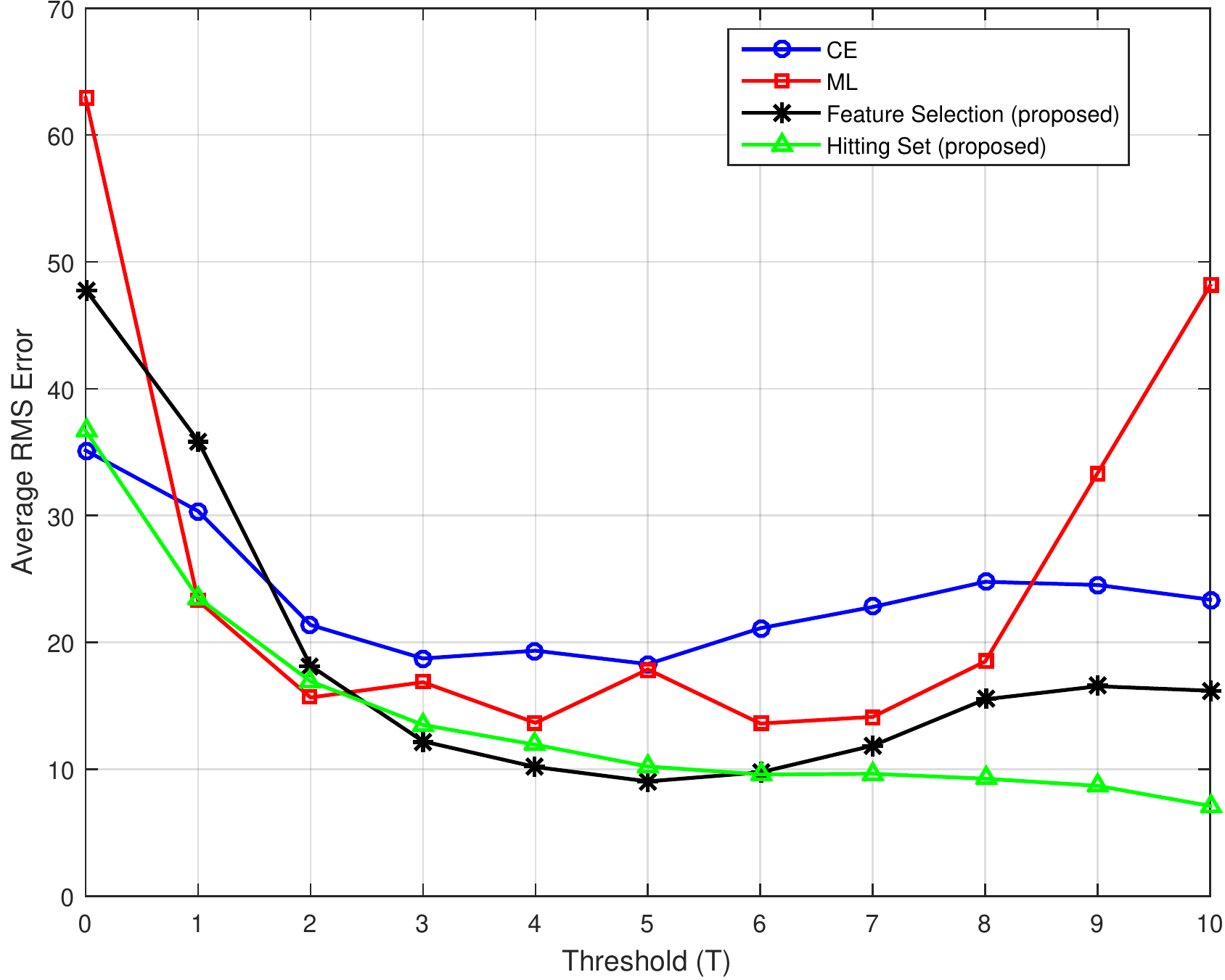}
  \caption{$M = 100, N = 200$}
  \label{n200}
\end{subfigure}
\caption{Source localization performance with different values of threshold T and fault probability $P_f = 0.1$.}
\label{threshold}
\end{figure*}

Fig. \ref{threshold} shows the performance of CE, ML, Feature Selection (Algorithm 1), and Hitting Set approach (Section 3.1) with different values of threshold T. It can be observed from Fig. \ref{n150} and \ref{n200} that threshold value $T = 5$ can be selected as an optimal value for different sample complexity (e.g. M = 50, 100) and number of sensor nodes (e.g., N = 150, 200) because it provides better performances for majority of estimators. Hence, we use threshold value $T = 5$ for rest of the experiments.

\subsubsection{Fault tolerance analysis of proposed estimators}
In this subsection, we evaluate the fault tolerant capabilities of proposed estimators in comparison to CE and ML. In addition, the measurement noise $w_{n, t}$ is distributed according to $\mathcal{N}(0, 1)$. Moreover, similar simulation results for small sample size $M$ are presented in \textbf{Appendix A}.

\begin{figure*}[ht]
\centering
\begin{subfigure}{.4\textwidth}
  \centering
  \includegraphics[width=2.5in]{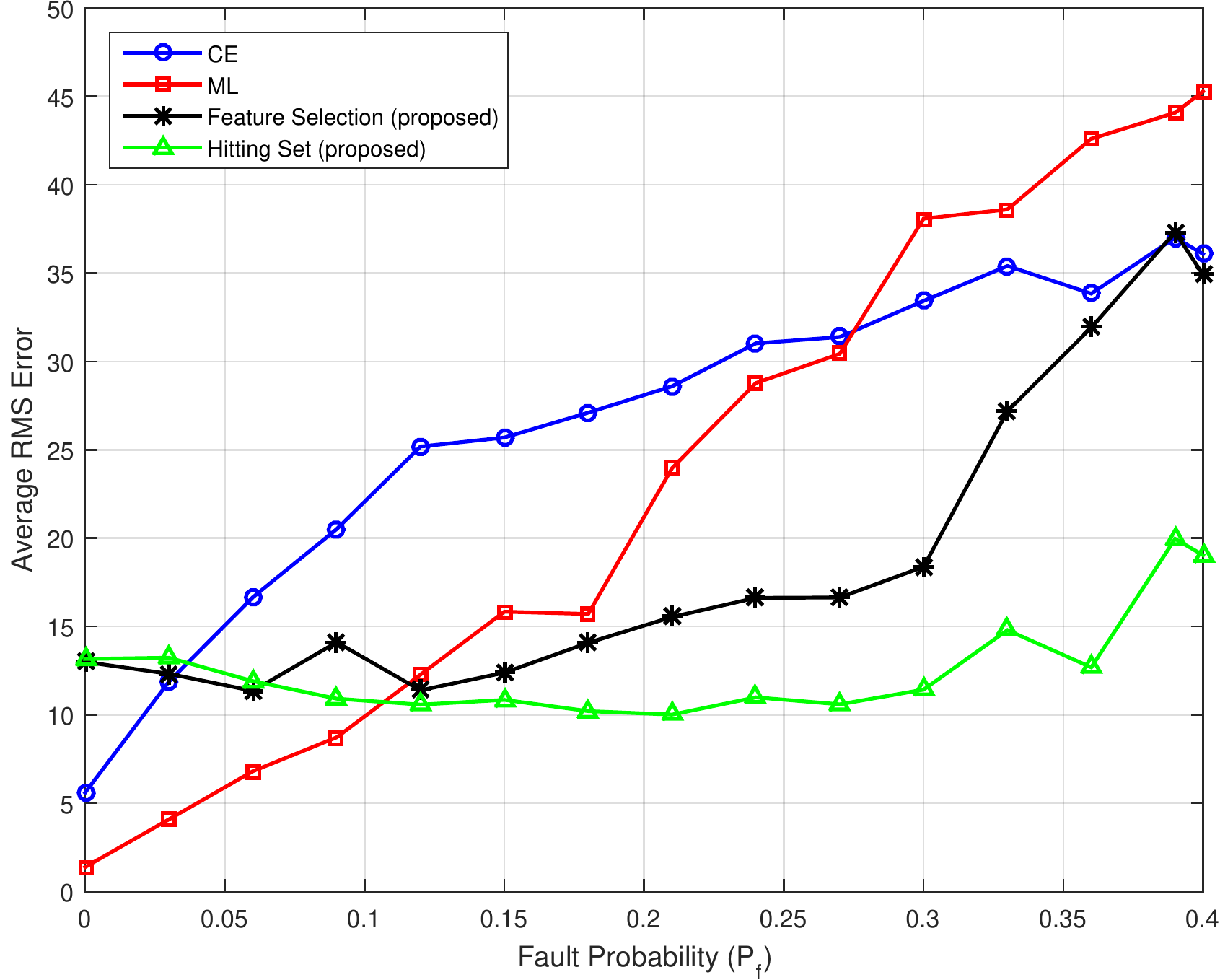}
  \caption{$M = 100, N = 150$}
  \label{mn150}
\end{subfigure}
\begin{subfigure}{.4\textwidth}
  \centering
  \includegraphics[width=2.5in]{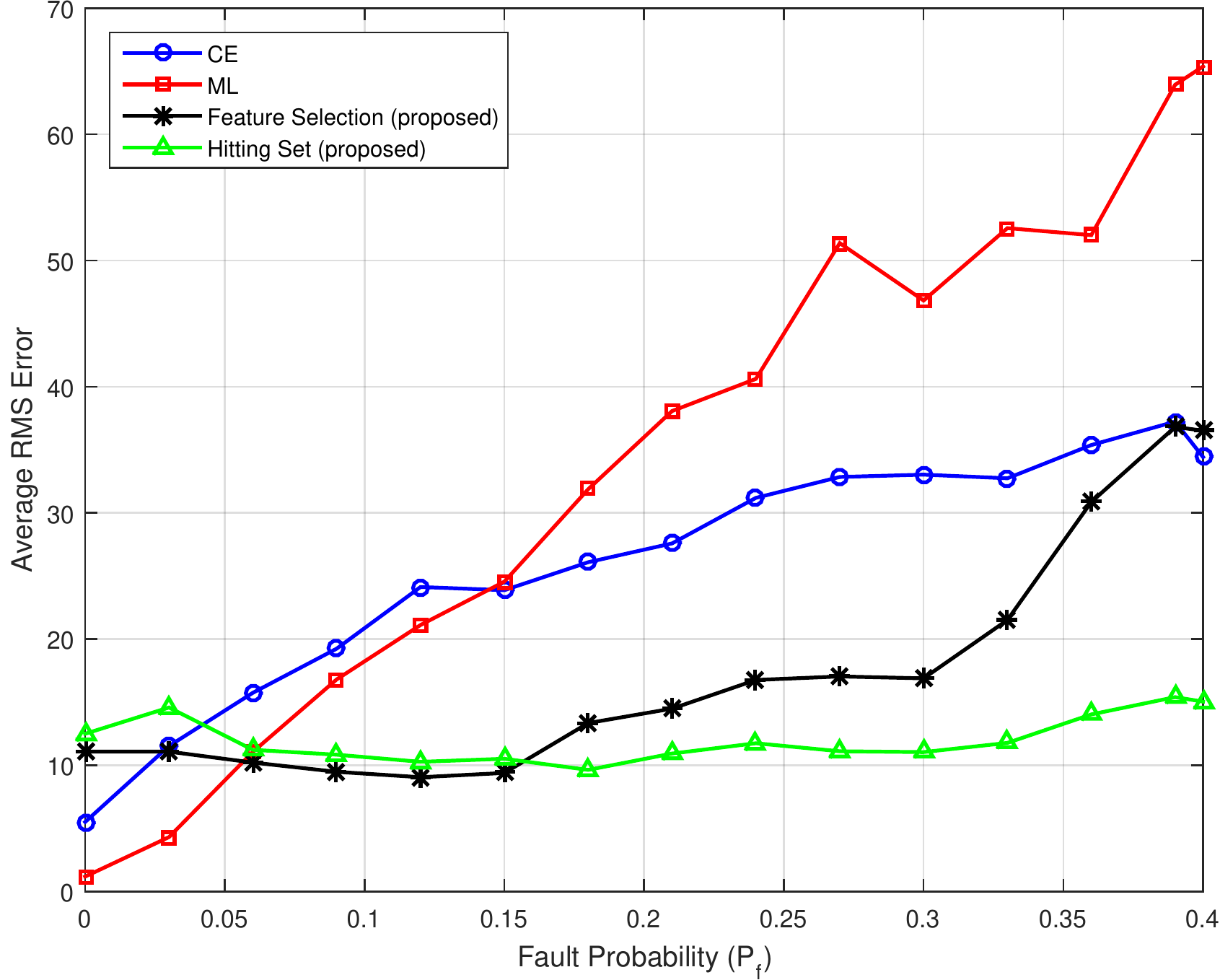}
  \caption{$M = 100, N = 200$}
  \label{mn200}
\end{subfigure}
\caption{Source localization performances of the proposed estimators vs. Maximum Likelihood (ML) or Centroid Estimator (CE).}
\label{faultprob}
\end{figure*}

The simulation results are shown in Fig. \ref{faultprob} for $M = 100$ samples and $N = 150, 200$ sensor nodes. Our proposed estimators (feature selection and hitting set approach) are more fault tolerant estimators than CE and ML as shown in Fig. \ref{mn150} and \ref{mn200} for only $M = 100$ samples. Moreover, hitting set approach achieves minimum RMS error for $N = 150, 200$ sensor nodes, and its performance improves by increasing the number of sensor nodes significantly especially for large fault probability because the source neighborhood estimation accuracy of hitting set improves by increasing sensor nodes. Similarly, feature selection method reduces RMS error significantly as compared to CE and ML, and it further enhances its performance especially for smaller fault probability when the number of sensor nodes increase. The performance of feature selection method degrades rapidly for a large fault probability such as $P_f = 0.3-0.4$ as shown in Fig. \ref{faultprob}.

On the other hand, CE and ML performances deteriorate about linearly as the fault probability increases as shown in Fig. \ref{faultprob} where CE estimated source location has almost no effect of increasing the number of sensor nodes while ML estimated source location gets far away from the true location as the sensor nodes increase. Moreover, ML is very sensitive to false negatives as compared to CE which is sensitive to false positives (but not as much as ML to false negatives), therefore, CE has better performance than ML for a large number of sensor nodes as shown in Fig. \ref{mn200}.

\subsubsection{Increasing sample complexity impact analysis}
\begin{figure}[t]
\centering
\includegraphics[width=2.5in]{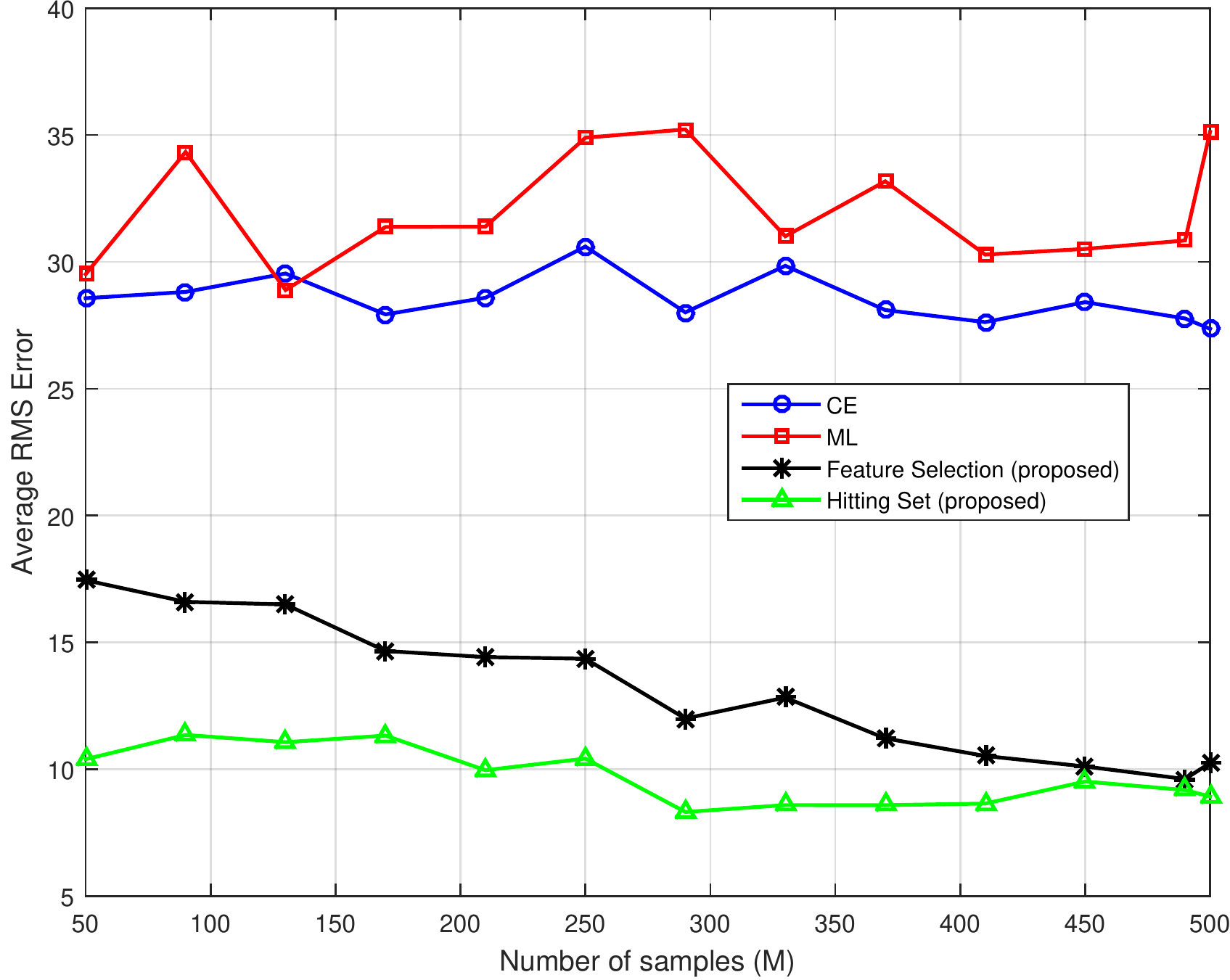}
\caption{Source localization performances of the proposed methods vs. ML or CE for increasing sample complexity $M$, $P_f = 0.2$, and $N = 200$.}
\label{samplecomp}
\end{figure}

We analyze the impact of sample complexity on the proposed estimators, CE, and ML in this subsection. The measurement noise is distributed according to standard Gaussian distribution. The fault probability $P_f = 0.2$ and the number of sensor nodes $N = 200$ are used for this experiment. The result is shown in Fig. \ref{samplecomp} where the RMS error decreases for both the proposed estimators by increasing the sample complexity $M$. However, the performance improvement is significant for the feature selection method as compared to hitting set approach because feature selection method improves its estimation of neighborhood of the source by increasing the sample size of the data set as shown in Algorithm 1. In addition, a very small improvement happens in CE performance while the performance of ML estimator seems invariable with respect to sample complexity.

\subsubsection{Impact of increasing sensor nodes}
Here we study the impact of increasing sensor nodes on the performances of proposed estimators, CE, and ML. The measurement noise is standard Gaussian, and fault probability and sample complexity are $P_f = 0.1$ and $M = 200$, respectively.

\begin{figure}[ht]
\centering
\includegraphics[width=2.5in]{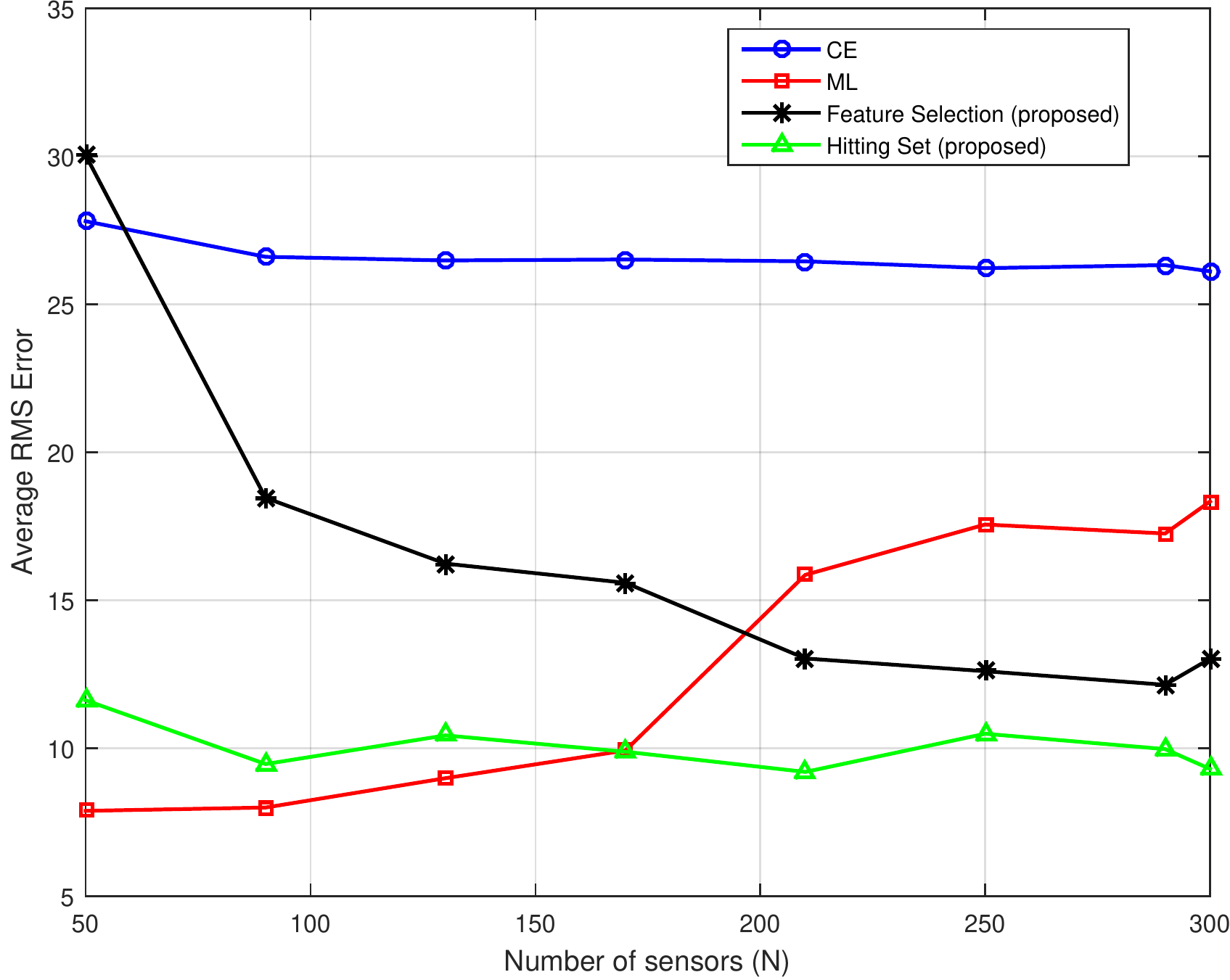}
\caption{Performance analysis of the proposed methods, ML, and CE with $P_f = 0.1$ and $M = 200$ samples for increasing sensor nodes $N$.}
\label{sennodes}
\end{figure}

Fig. \ref{sennodes} shows the simulation result where proposed estimators (feature selection and hitting set methods) are sensitive to the number of sensor nodes such that their performances improve by increasing the number of sensor nodes. Moreover, the feature selection method performance gets better significantly by increasing the number of sensor nodes while hitting set method's performance improvement is not sufficient enough for the large sample size $M = 200$.

On the other hand, CE is invariable to increase in sensor nodes while ML performance decreases rapidly with the increase in sensor nodes as shown in Fig. \ref{sennodes}. Moreover, similar phenomena is depicted by Fig. \ref{faultprob} for ML. The reason for this rapid reduction in performance of ML with respect to the number of sensor nodes is due to the fact that ML is very sensitive to false negatives. Therefore, more sensor nodes will increase the probability of false negatives.

\subsubsection{Impact of measurement noise variance}
\begin{figure*}[ht]
\centering
\begin{subfigure}{.4\textwidth}
  \centering
  \includegraphics[width=2.5in]{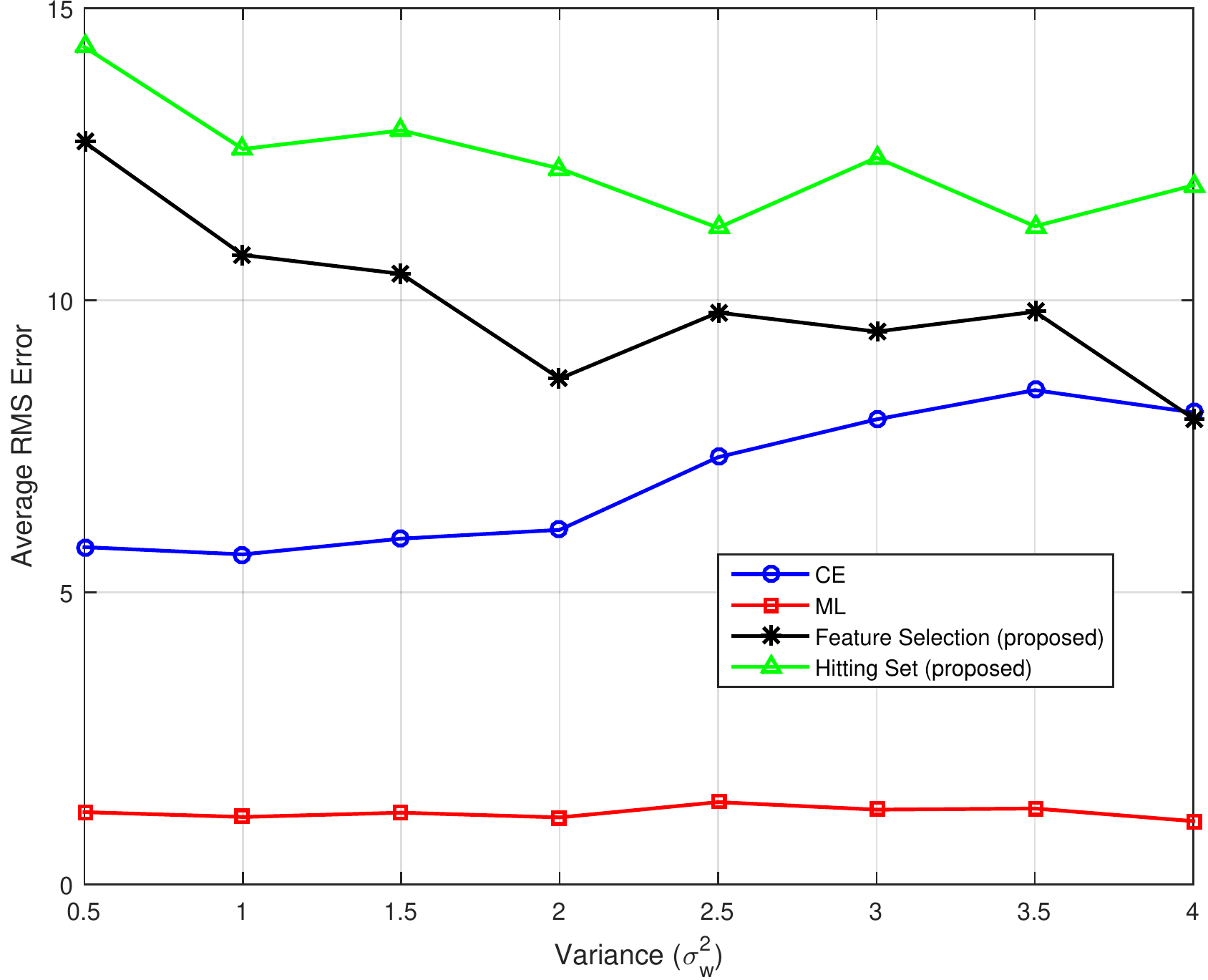}
  \caption{$P_f = 0$, $M = 50$, $N = 200$}
  \label{vp00}
\end{subfigure}
\begin{subfigure}{.4\textwidth}
  \centering
  \includegraphics[width=2.5in]{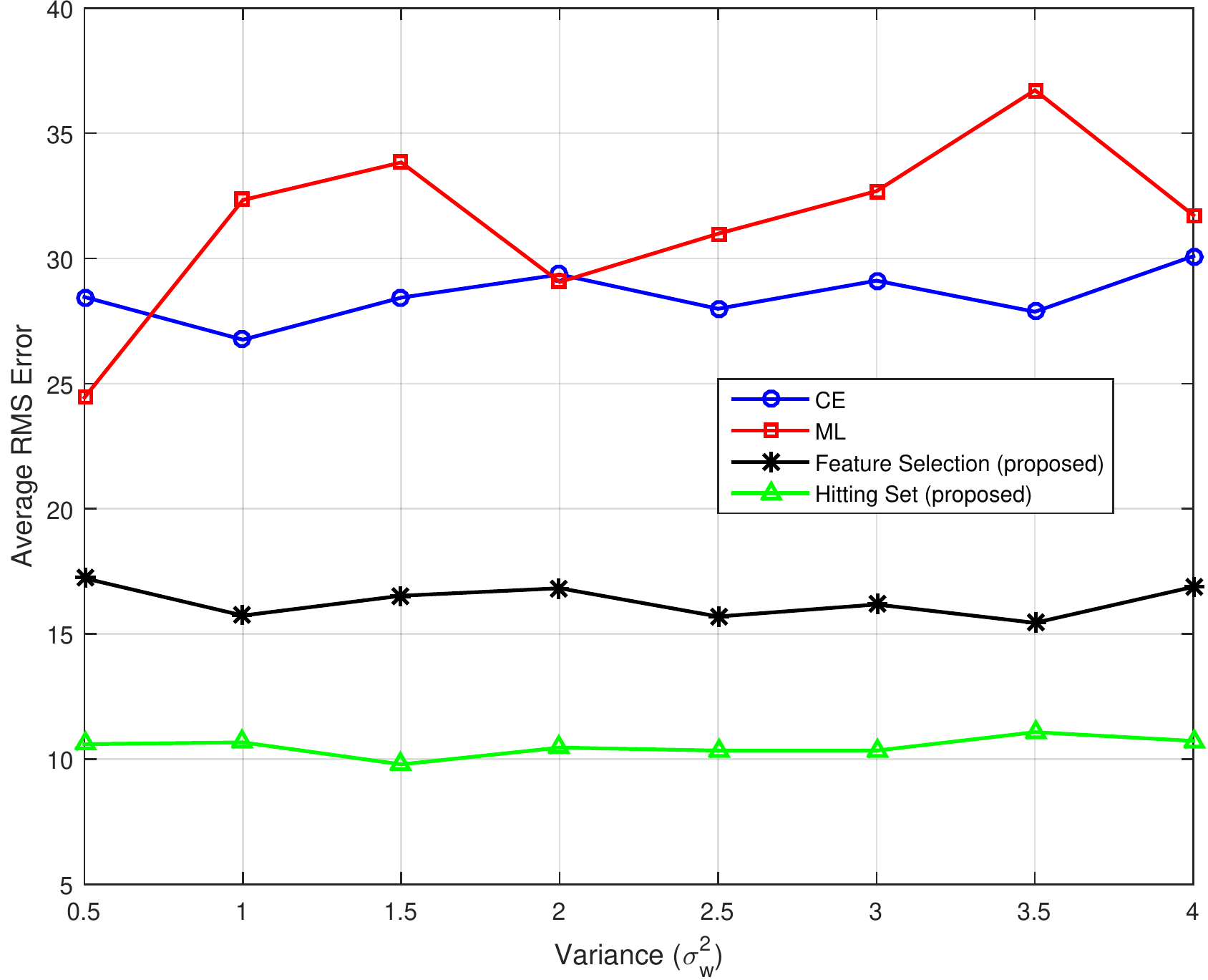}
  \caption{$P_f = 0.2$, $M = 50$, $N = 200$}
  \label{vp02}
\end{subfigure}
\caption{One source localization performances of the proposed estimators, ML, and CE with increase in the measurement noise variance $\sigma_w^2$.}
\label{var}
\end{figure*}

We evaluate the performances of proposed estimators, CE, and ML by increasing the Gaussian noise variance $\sigma_w^2$. The results are shown in Fig. \ref{var} with fault probability $P_f = 0$ and $P_f = 0.2$, sample complexity $M = 50$, and the number of sensor nodes $N = 200$.

When there is no fault (false positive or false negative), ML is the best source localization estimator. Moreover, the performance of ML does not degrade with measurement noise ($\sigma_w^2$) as shown in Fig. \ref{vp00}. On the other hand, CE performance decreases significantly as the variance increases. While the proposed estimators provide worst performances because these estimators are fault tolerant source localization estimators, and they achieve the best results when fault probability is nonzero as shown in Fig. \ref{vp02}. When fault probability is $P_f = 0.2$, all the estimators become invariable to increase in noise variance as shown in Fig. \ref{vp02} because all the estimators are sensitive to fault probability not much to measurement noise variance. In addition, they are not much sensitive to measurement noise due to enough sample complexity M as well.

\subsubsection{Results summary of Fig. \ref{faultprob} to Fig. \ref{var}}
We summarize the simulation results of Fig. \ref{faultprob} to Fig. \ref{var} here. All the estimators including proposed estimators studied in this paper are sensitive to fault probability $P_f$. However, the proposed estimators (feature selection and hitting set methods) are more fault tolerant than CE and ML. In addition, the hitting set approach is the best source localization method especially for the fault model with a large fault probability.

First of all the centroid estimator (CE) is almost invariable to increase in sensor nodes while there is a small positive effect of sample complexity on it. Moreover, CE is sensitive to measurement noise ($\sigma_w^2$) such that its performance reduces with increase in measurement noise when fault probability is zero.

The maximum likelihood (ML) estimator is very sensitive to increase in sensor node because more sensors lead to more false negatives, hence, decreases its performance. On the other hand, it is unchanged with respect to sample complexity and measurement noise.

Finally the proposed estimators (feature selection and hitting set methods) improve their performance with increase in sample complexity or the number of sensor nodes which are the fundamental characteristics of any estimators. They do not change with respect to measurement noise ($\sigma_w^2$) (insensitive to measurement noise). Furthermore, the hitting set approach provides the best results in the simulation results of Fig. \ref{faultprob} to Fig. \ref{var} with nonzero fault model probability.

\subsection{Comparison with FTML and SNAP Algorithm for a Single Source}
In this section, we compare our estimator (feature selection approach) with FTML \cite{MP} and SNAP algorithm \cite{Michaelides} for a single source localization. We have pointed out in Section 2.4 that FTML and SNAP algorithm take into account the knowledge of fault probability $P_f$ and knowledge of ROI, respectively, which are not easy to obtain (or meet accurately) in general. Therefore, we also perform simulations when only the rough estimates of fault probability $P_f$ and Region of Influence (ROI) of the source are available to FTML and SNAP algorithm, respectively in \textbf{Appendix B}.

We modify our feature selection approach to employ maximum likelihood (ML) formula (\ref{eq:5}) after selecting more relevant sensor nodes to improve its performance with the cost of increased complexity.
\begin{figure*}[ht]
\centering
\begin{subfigure}{.4\textwidth}
  \centering
  \includegraphics[width=2.5in]{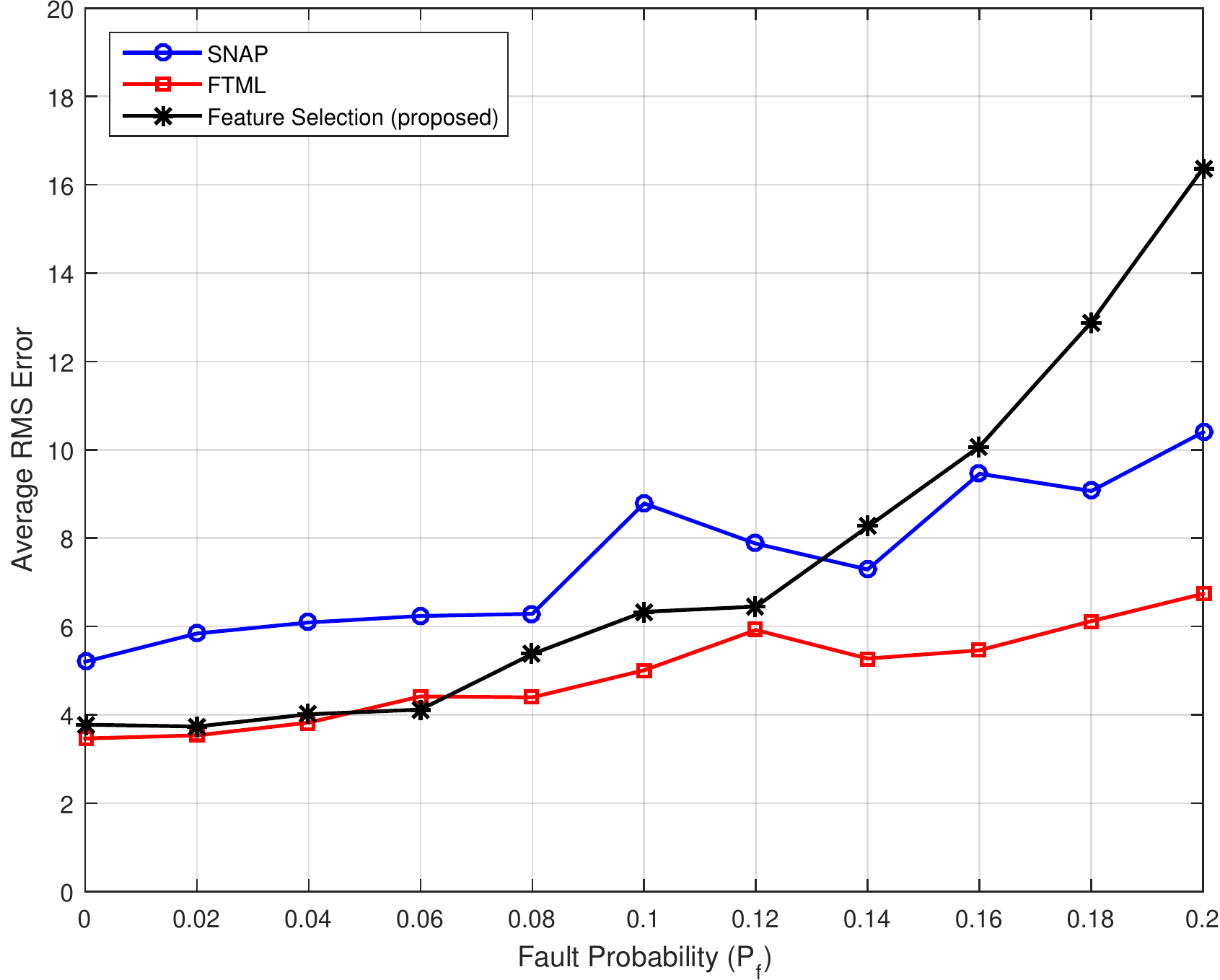}
  \caption{$M = 20$, $N = 50$}
  \label{ML1}
\end{subfigure}
\begin{subfigure}{.4\textwidth}
  \centering
  \includegraphics[width=2.5in]{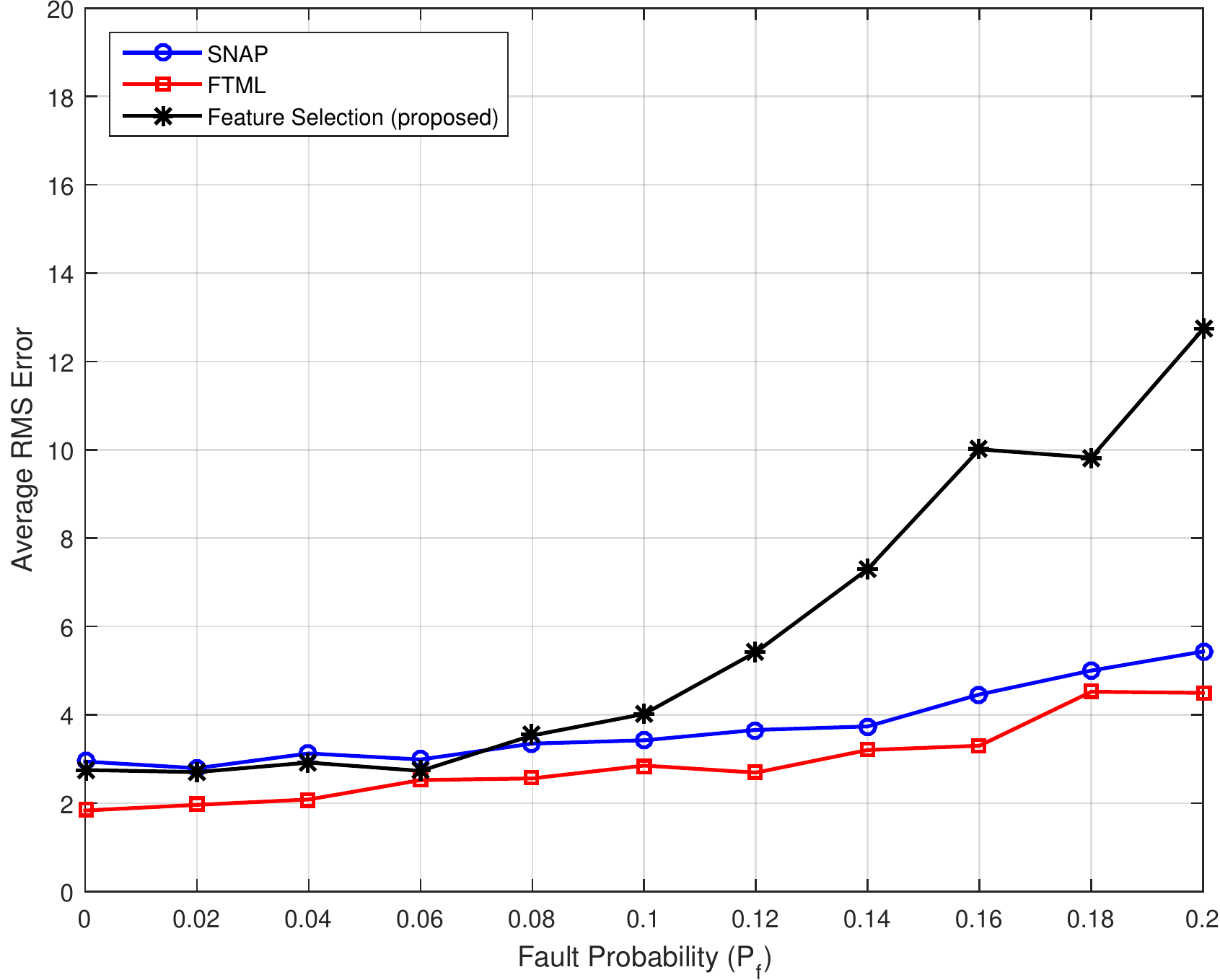}
  \caption{$M = 20$, $N = 100$}
  \label{ML2}
\end{subfigure}
\caption{One source localization performance with modified feature selection approach vs FTML with perfect knowledge of $P_f$ or SNAP with perfect knowledge of ROI.}
\label{MLmod}
\end{figure*}

The result is shown in Fig. \ref{MLmod} with $M = 20$ samples and $N = 50, 100$ sensor nodes with standard Gaussian measurement noise $w_{n,t}$. It can be observed from Fig. \ref{ML1} that our feature selection approach performs better than FTML \cite{MP} and SNAP algorithm \cite{Michaelides} for small fault probability $P_f$ and $N = 50$ sensor nodes when modified to use ML estimator. By increasing the sensor nodes, the performance also improves for all the estimators (e.g., proposed feature selection, FTML, SNAP), however, the increase in performance is significant for FTML and SNAP algorithm as shown in Fig. \ref{ML2}. The advantage of our approach is that it works with noisy data set without any knowledge of $P_f$ and region of influence (ROI).

\subsection{Simulation for Two Sources}
In this subsection, we simulate only our proposed techniques for two sources because we expect them to have better performances as compared to Centroid Estimator and ML Estimator as for a single source. We estimate the clusters of sensor nodes once at the start of the simulations. However, we can also employ the iterative method as proposed in Algorithm 2 for iteratively estimating the clusters and sources' locations. After estimating the clusters of sensor nodes, we use the hitting set/feature selection approach separately for each cluster as proposed in Section 4.

For the simulations, we utilize almost the same setting as used for a single source where observation model in formula (\ref{eq:11}) for $K = 2$ and average RMS error in formula (\ref{eq:12}) are used for data set and the performance metric, respectively. In addition, the value of threshold $T = 5$ is chosen because it achieved better performances for a single source, and the measurement noise variance is standard Gaussian random variable.

\subsubsection{Fault tolerance analysis for two sources} To study the fault tolerance of the proposed estimators for two sources with uniformly random locations, we perform simulation for $N = 200$ numbers of sensor nodes and $M = 50$ samples.

\begin{figure}[ht]
\centering
\includegraphics[width=2.5in]{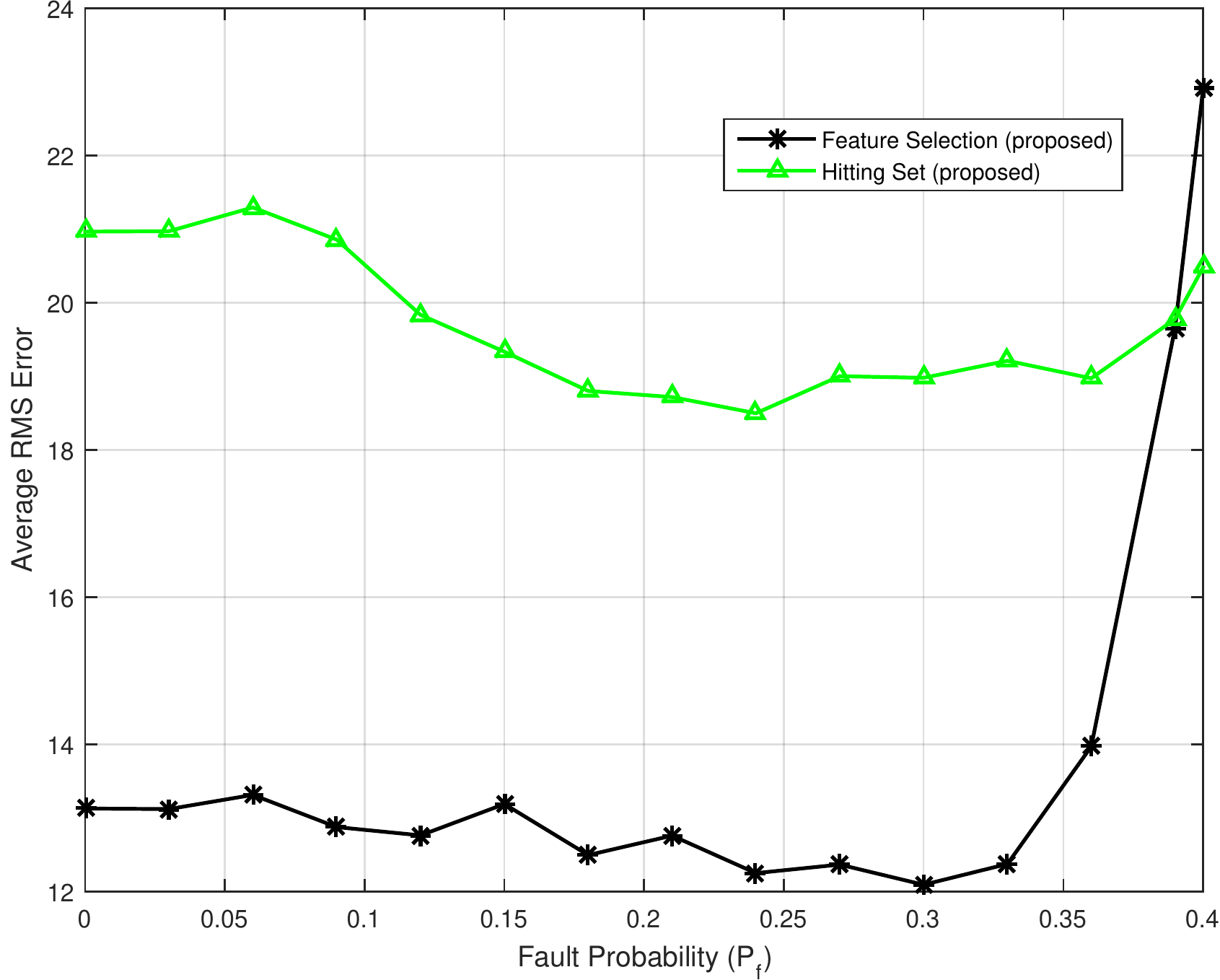}
\caption{Two sources localization performances of the proposed estimators with uniformly random locations vs. the fault probability $P_f$ for $M = 50$ and $N = 200$.}
\label{twosources}
\end{figure}

The result is shown in Fig. \ref{twosources} where the performances of both the proposed estimators (hitting set and feature selection) are not affected much for the fault probability in the range $[0, 0.3]$ because of average performances. On the other hand, significant performance degradation happens for both the estimators for the fault probability in the range $[0.3, 0.4]$, however, feature selection performance deteriorates sharply. For two sources, the feature selection method performs better than the hitting set approach which suggests that the hitting set approach is sensitive to the initial estimates of the clusters of sensor nodes.

\subsubsection{Performance analysis for the distance between two sources} We also study the impact of distance between two sources on the performances where the distance is measured by $L_2$ norm. Furthermore, the fault probability is set to be $P_f = 0.2$, and the number of samples $M = 50$ and the number of sensor nodes $N = 200$ are selected for the simulation.

\begin{figure}[ht]
\centering
\includegraphics[width=2.5in]{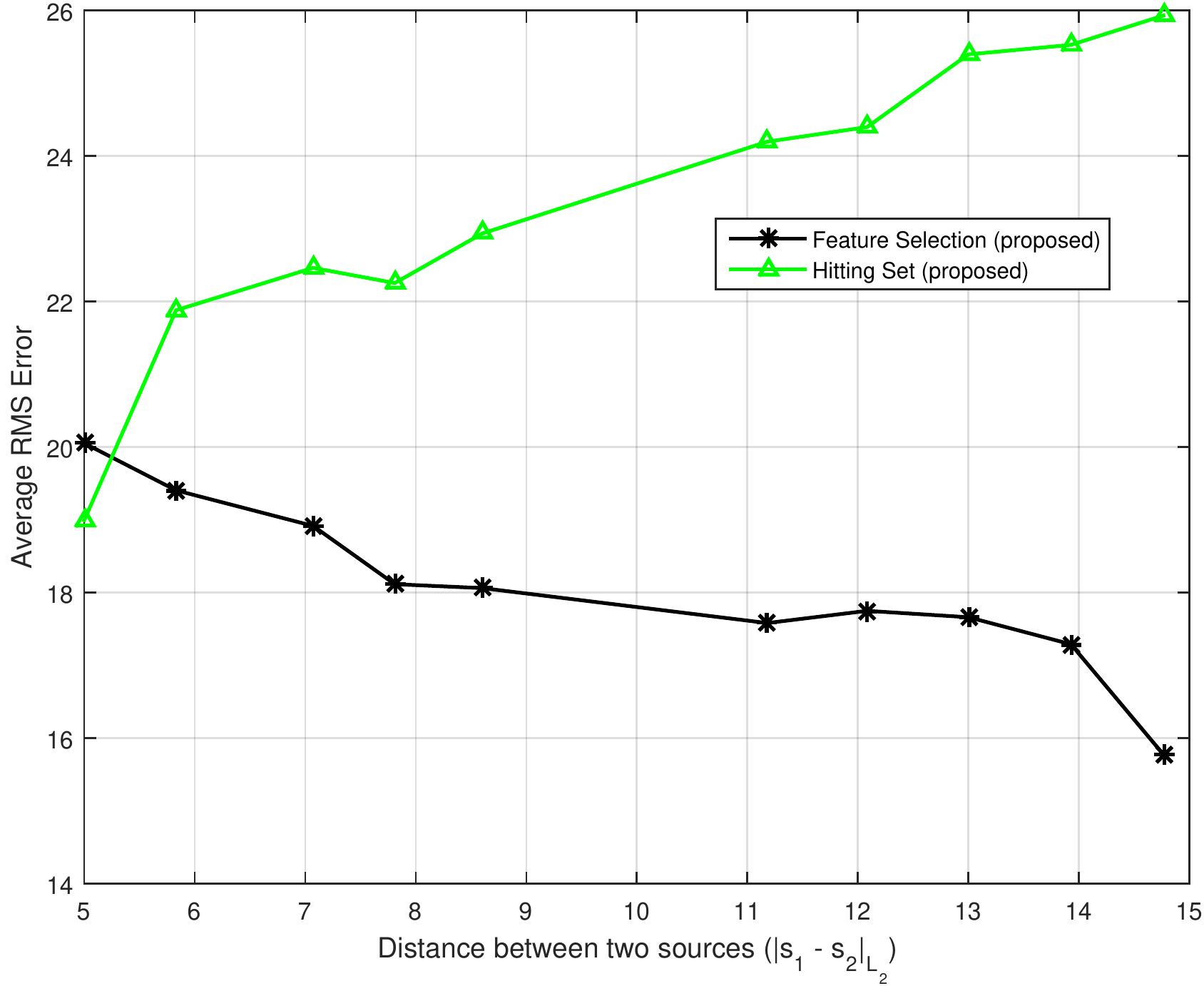}
\caption{Two sources localization performances of the proposed estimators vs distance between the sources for $P_f = 0.2$, $M = 50$, and $N = 200$.}
\label{distancetwo}
\end{figure}

The result is shown in Fig. \ref{distancetwo} where the opposite effect of the distance between two sources can be noticed on the two estimators. As the distance between two sources increases, the performance of feature selection gets better. On the other hand, the performance of hitting set approach worsens as the distance gets larger. The phenomenon of Fig. \ref{distancetwo} can be explained as follows:
\begin{enumerate}
  \item The feature selection approach is proposed in Algorithm 1 where it can be observed that the most relevant columns (i.e., sensor nodes) of the data set are selected for estimating the source location. As the distance between two sources increases, it becomes convenient to select the most relevant sensor nodes for each source because the least relevant sensor nodes get alarmed rarely. Hence the performance improves with the increase in distance between the two sources.
  \item In the hitting set approach, we select the alarmed nodes for each sample (collection for whole data set). Then the hitting set is estimated from the collection. The increase in distance between the two sources, and then the initial estimation of two clusters of sensor nodes could include the wrong sensor nodes in wrong estimated clusters of sensor nodes. This can happen for the sensor nodes which are located at the boundaries of ROIs of two sources. The sensor nodes, which are put falsely in the estimated clusters, are included in the collections, and then selected for each source localization in the hitting set approach. This degrades the performance of hitting set approach for two sources localization. However, incorrect estimated clusters have minimum impact on the feature selection approach because the sensor nodes, which are included in the wrong clusters, get alarm rarely for the false source.
\end{enumerate}

\section{Conclusion}
We studied the decentralized source localization problem under the fault model in this paper, and proposed two approaches: hitting set and feature selection, which are based on the noisy data set. The theoretical properties of these approaches have been analyzed and proved to be more fault tolerant comparatively. We also derived a lower bound on the sample complexity for hitting set approach. In addition, the extensions of the proposed methods for multiple sources localization in static and iterative settings were proposed as well. The proposed methods had been validated by extensive simulations for various parameters such as sample complexity, the number of sensor nodes, variance of the noise, and so on. It has been observed that the proposed estimators achieved the better performance compared to centroid and maximum likelihood estimators in a single source localization setting, and comparable performances with respect to FTML and SNAP algorithm in special settings. For the two sources case, the feature selection approach has better performance because it is insensitive to the initial estimates of the clusters of sensor nodes. Further studies can be carried out by extending the proposed methods in other directions such as adaptive quantization, source tracking, and packet loss problem.

\appendices
\section{Simulation results for small sample size M}
\begin{figure}[ht]
\centering
\begin{subfigure}{.4\textwidth}
  \centering
  \includegraphics[width=2.5in]{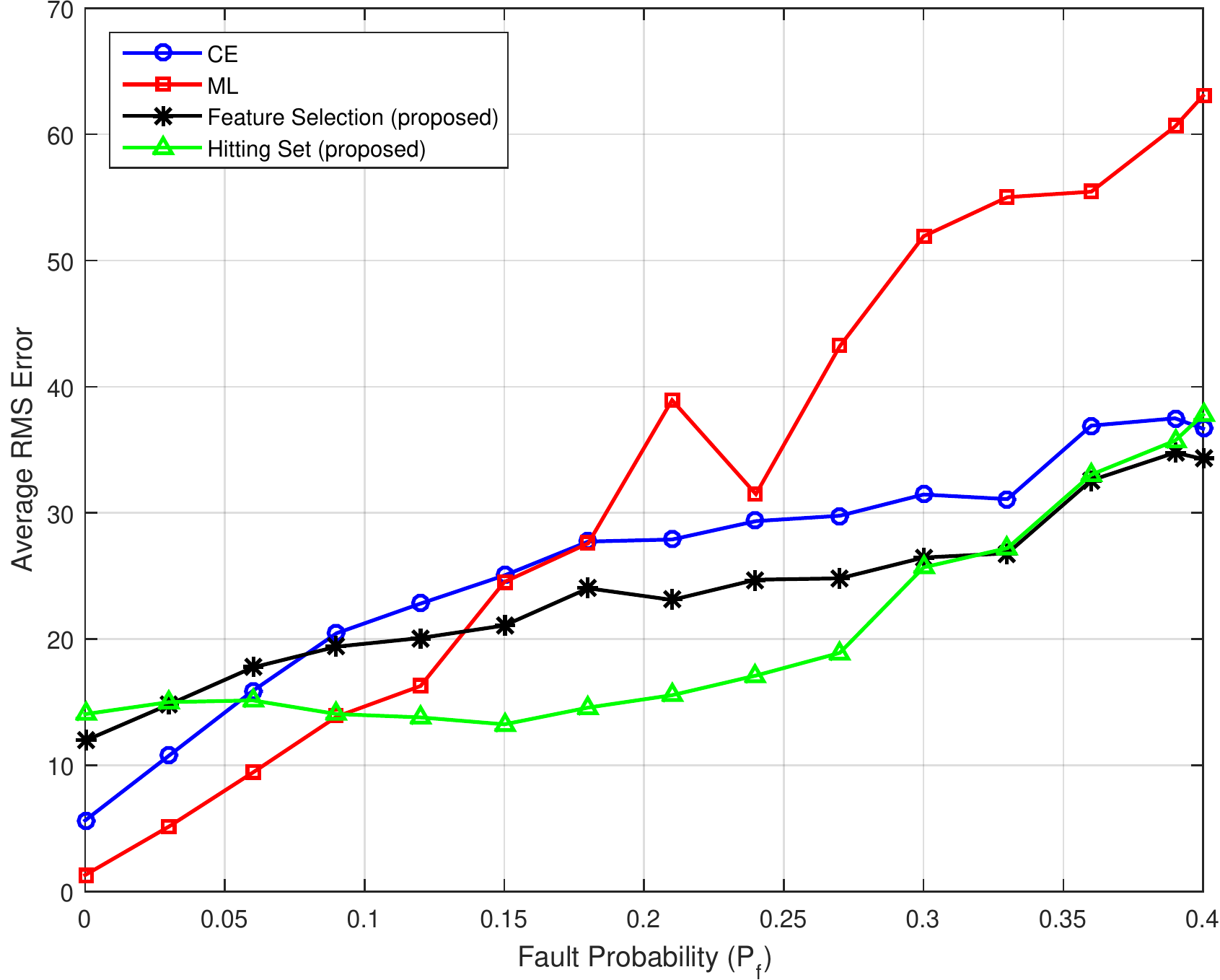}
  \caption{$M = 10$, $N = 200$}
  \label{M10}
\end{subfigure}
\begin{subfigure}{.4\textwidth}
  \centering
  \includegraphics[width=2.5in]{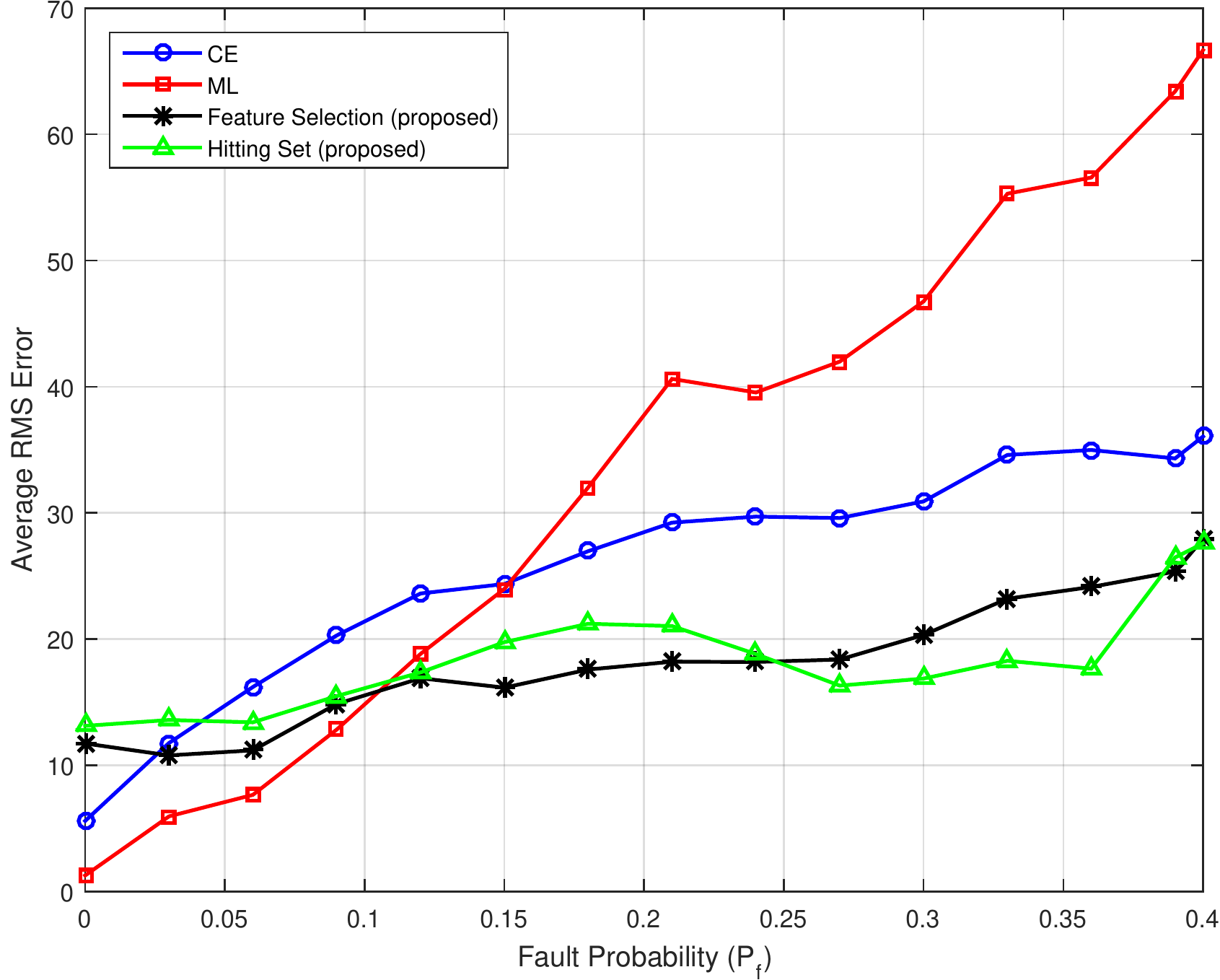}
  \caption{$M = 20$, $N = 200$}
  \label{M20}
\end{subfigure}
\caption{One source localization performances of the proposed estimators vs. ML or CE for small sample size $M$.}
\label{append}
\end{figure}

In this section, we study the fault tolerant performances of the proposed estimators in addition of maximum likelihood (ML) estimator and centroid estimator (CE) for small sample size $M$. Following the same setting as for Fig. \ref{faultprob}, we simulate the results for $M = 10, 20$ samples and $N = 200$ number of sensor nodes as shown in Fig. \ref{append}.

It can be observed that the proposed estimators (hitting set and feature selection) are more fault tolerant as compared to ML and CE estimators even for small sample size $M$. Furthermore, the hitting set approach performs better than the feature selection approach for $M = 10$ samples as shown in Fig. \ref{M10}. However, the feature selection performance is better than the hitting set approach for $M = 20$ samples especially for fault probability in the range $[0, 0.25]$ as shown in Fig. \ref{M20} because the feature selection approach is highly data dependent. It requires more samples to perform better as shown in Algorithm 1.

In addition, we have noticed that the performance of our proposed feature selection approach in Fig. \ref{M20} (for $M = 20$ samples) is better than Fig. \ref{mn200} (for $M = 100$ samples) for fault probability in the range $[0.35, 0.4]$. This better performance with small sample size $M$ with high probability of fault model may be due the following reasons: 1) the weight vector $W = [1, 1, 1]$ is more optimal to feature selection approach for small sample size, 2) the feature selection criteria (e.g., line 3-11 of Algorithm 1) is more suitable for small sample with high probability of fault model.

\section{Comparison with FTML and SNAP with imperfect knowledge of the fault model probability and ROI}
Generally the perfect knowledge of $P_f$ and region of influence (ROI) of the source is not available, therefore, in this appendix, we compare our estimators with FTML estimator \cite{MP} and SNAP algorithm \cite{Michaelides} when only the noisy estimates of fault model probability $P_f$ and ROI can be obtained. We have discussed that the proposed estimators (hitting set method and feature selection approach) do not require these estimates while FTML and SNAP algorithm require them.

In the following results, the fault model probability $P_f$ and the knowledge of ROI are made noisy as follows: $P_f + 0.05$ for FTML estimator and $1.2\times ROI$ for SNAP algorithm. In addition, we use $M = 20$ samples and $N = 50, 100$ sensor nodes with standard Gaussian measurement noise $w_{n,t}$.

\subsection{Without any modification}
\begin{figure}[ht]
\centering
\includegraphics[width=2.5in]{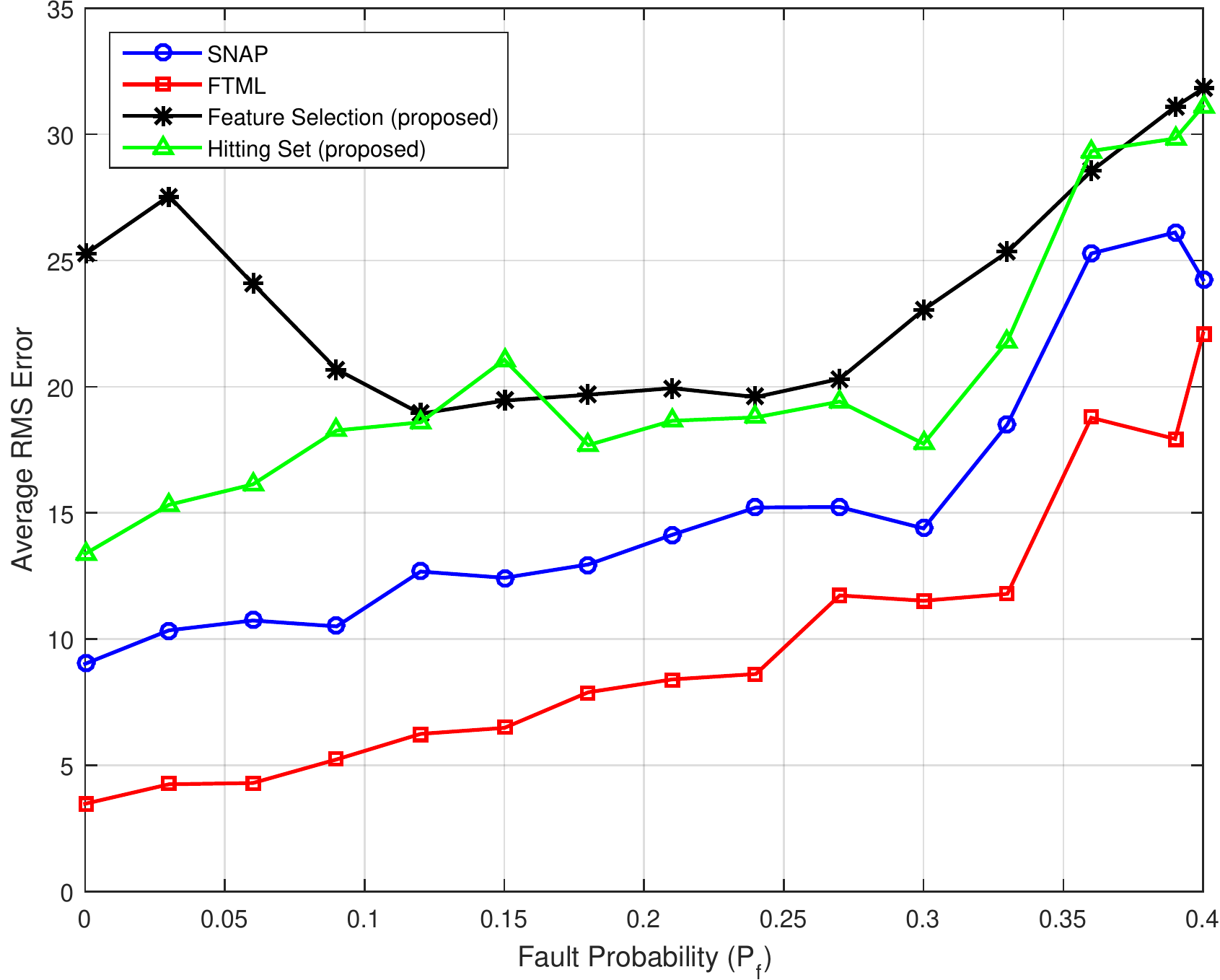}
\caption{One source localization performance without any modification to proposed estimators vs. FTML with slightly imperfect knowledge of $P_f$ or SNAP with slightly imperfect knowledge of ROI for $M = 20$ and $N = 50$.}
\label{CEM}
\end{figure}

The simulation result is shown in Fig. \ref{CEM} which is obtained without any modification to the proposed estimators. Overall, the hitting set approach and SNAP algorithm with distorted ROI are almost comparable for the range of fault model probability as shown in Fig. \ref{CEM}. While FTML estimator has the best performance even with noisy estimate of fault model probability (e.g., $P_f + 0.05$).

\subsection{With modified feature selection approach}
\begin{figure}[ht]
\centering
\begin{subfigure}{.4\textwidth}
  \centering
  \includegraphics[width=2.5in]{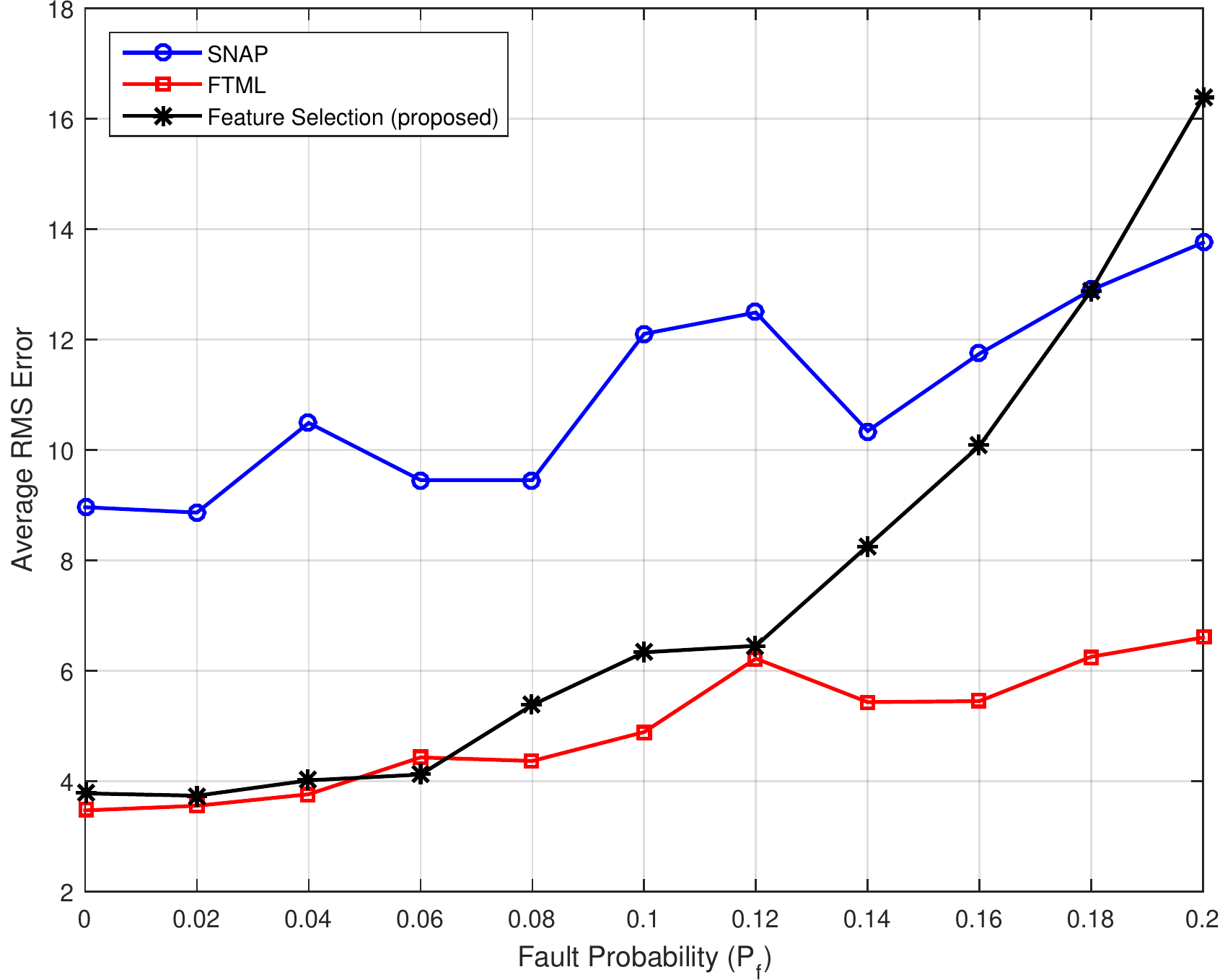}
  \caption{$M = 20$, $N = 50$}
  \label{DML1}
\end{subfigure}
\begin{subfigure}{.4\textwidth}
  \centering
  \includegraphics[width=2.5in]{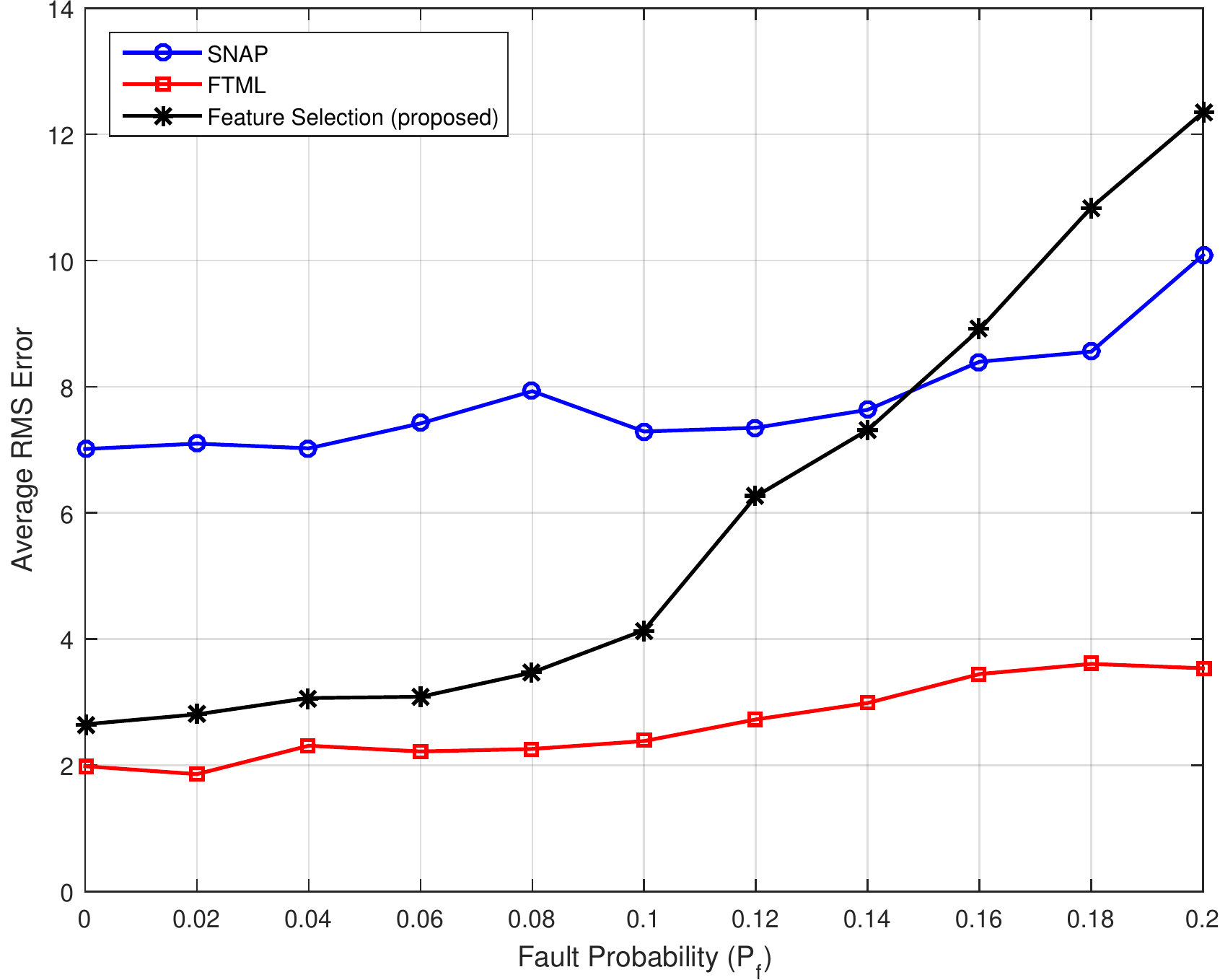}
  \caption{$M = 20$, $N = 100$}
  \label{DML2}
\end{subfigure}
\caption{One source localization performance with modified feature selection approach vs. FTML with slightly imperfect knowledge of $P_f$ or SNAP with slightly imperfect knowledge of ROI.}
\label{DML}
\end{figure}

The feature selection approach is modified to use maximum likelihood (ML) estimator after selecting the most relevant sensor nodes (i.e., alarmed at least $25\%$ of the number of samples). The performance of the proposed estimator is improved with the cost of complexity, however, the final step is performed by the fusion center (FC). Fig. \ref{DML} shows the simulation result in comparison to FTML and SNAP algorithm with imperfect knowledge of the fault probability and ROI with $M = 20$ samples and $N = 50$ and $N = 100$ sensor nodes. It can be observed from the result that the feature selection approach performs comparable to FTML and much better than SNAP algorithm for small fault model probability $P_f$. Moreover, increasing the sensor nodes has the positive impact on the source localization performance of all the estimators. We have mentioned previously that the feature selection approach needs only the noisy data set for source localization.

\section*{Acknowledgments}
We would like to thank Dr. Michalis P. Michaelides for providing us the codes of his paper \cite{MP}, and reviewing the draft of this paper.




\end{document}